\documentclass{article}

\input{glyphtounicode}  
\usepackage[colorlinks,pagebackref,linktocpage]{hyperref}

\usepackage{amssymb}  
\usepackage{amsmath}   

\usepackage{tablefootnote} 

\usepackage{amsthm} 

\usepackage{amsopn}

\usepackage{color}
\definecolor{orange}{RGB}{255,127,0}
\definecolor{brown}{RGB}{150,70,0}
\definecolor{green}{RGB}{127,255,127}
\definecolor{darkgreen}{RGB}{0,127,0}
\definecolor{blue}{RGB}{127,127,255}
\definecolor{lightblue}{RGB}{150,150,255}
\definecolor{darkblue}{RGB}{0,0,127}
\definecolor{red}{RGB}{255,90,90}
\definecolor{grey}{RGB}{127,127,127}
\definecolor{pink}{RGB}{255,180,180}

\newcommand{\arxiv}[1]{{{#1}}}
\newcommand{\journal}[1]{\textcolor{darkblue}{{}}}

\newcommand{\comment}[1]{}

\usepackage{graphicx}

\title{ \journal{Evaluation in AI: \\ From task-oriented to ability-oriented measurement}  
       \arxiv{AI Evaluation: past, present and future\footnote{This paper corresponds to a lecture given for the Summer School of the Spanish Association for Artificial Intelligence, in A Coru\~na, Spain, September 2014.}} 
}			
\author
{
	Jos\'{e} Hern\'{a}ndez-Orallo\\
	{\normalsize\em DSIC, Universitat Polit\`ecnica de Val\`encia, Spain}\\
	{\normalsize \tt jorallo@dsic.upv.es}
}
\date{\today}

\begin{document}

\maketitle

\noindent {\bf This paper is largely superseded by the following paper:\\
``Evaluation in artificial intelligence: from task-oriented to ability-oriented measurement'' Journal of Artificial Intelligence Review (2016). doi:10.1007/s10462-016-9505-7, \url{http://dx.doi.org/10.1007/s10462-016-9505-7}.\\
Please check and refer to the journal paper.}

\begin{abstract}\arxiv{Artificial intelligence develops techniques and systems whose performance must be evaluated on a regular basis in order to certify and foster progress in the discipline. We will describe and critically assess the different ways AI systems are evaluated. We first focus on the traditional {\em task}-oriented evaluation approach. We see that black-box (behavioural evaluation) is becoming more and more common, as AI systems are becoming more complex and unpredictable. We identify three kinds of evaluation: human discrimination, problem benchmarks and peer confrontation. We describe the limitations of the many evaluation settings and competitions in these three categories and propose several ideas for a more systematic and robust evaluation. We then focus on a less customary (and challenging) {\em ability}-oriented evaluation approach, where a system is characterised by its (cognitive) abilities, rather than by the tasks it is designed to solve. We discuss several possibilities: the adaptation of cognitive tests used for humans and animals, the development of tests derived from algorithmic information theory or more general approaches under the perspective of universal psychometrics.}\vspace{0.2cm}
\end{abstract}

\section{Introduction}

The evaluation of any discipline must necessarily be linked to the purpose of the discipline. What is the purpose of artificial intelligence (AI)? McCarthy's pristine definition of AI sets this unambiguously: ``[AI is] the science and engineering of making intelligent machines'' \cite{McCarthy2007}. As a consequence, AI evaluation should focus on evaluating the intelligence of the artefacts it builds. However, as we will further discuss below, `intelligence tests' (of whatever kind) are not the everyday evaluation approach for AI. The explanation for this is that most AI research is better identified by Minsky's more pragmatic definition: 
``[AI is] the science of making machines capable of performing tasks that would require intelligence if done by [humans]''  
\cite[p.v]{Minsky1968}.  
As a result, AI evaluation focusses on checking whether machines do these {\em tasks} well.



This has led to an important anomaly of AI. AI artefacts solve these tasks {\em without featuring intelligence}. Paradoxically, this is one of the reasons of AI success. Systems are designed for a particular functionality and perform their task more {\em predictably} than humans, from driving cars to supply chain planning. Frequently, some tasks are not considered AI problems any more, once they are solved without full-fledged intelligence. This phenomenon is known as the ``AI effect'' \cite{McCorduck2004}. It would be unfair, however, to deny that some current AI systems, especially those that incorporate some learning potential, exhibit some intelligent behaviour. 

Anyway, it is not the purpose of this paper to dig further into the time-worn debate between narrow AI vs. general AI. Both approaches are valid and genuine parts of AI research. It is useful to have specialised AI systems that solve specific tasks, as well as systems that have abilities so that they can solve new problems they have never faced before. The intention of stressing this duality is that this should necessarily pervade the evaluation procedures in AI. Specialised AI systems should require a task-oriented evaluation, while general AI systems should require an ability-oriented evaluation.

This paper pays attention to the way evaluation is done in AI. As any science and engineering discipline, measuring is crucial for AI. 
Disciplines progress when they have objective evaluation tools to measure the elements and objects of study, assess the prototypes and artefacts that are being built and examine the discipline as a whole. As we will discuss in subsequent sections, despite the significant progress in the past couple of decades (with the generalisation of several AI benchmarks and competitions) there is still a huge margin of improvement in the way AI systems are evaluated. This is partially because we do not see AI evaluation as a {\em measurement process}\cite{hand2004measurement}. Also, it is probably a crucial moment to overhaul the way AI evaluation is performed, after the recent progress in areas of AI that are detaching from the narrow AI approach, such as developmental robotics \cite{Asada-etal2009}, deep learning \cite{Arel-etal2010}, inductive programming \arxiv{\cite{HernandezOrallo2014deep,Gulwani-etal2014dagstuhl}}\cite{Gulwani-etal2014cacm}, artificial general intelligence \cite{Goertzel-Pennachin2007}, universal artificial intelligence \cite{hutter2007universal}, etc. 

By overhauling AI evaluation, we aim at filling a gap, because, to our knowledge, there is no comprehensive analysis about how evaluation is performed in AI and how it can be improved and adapted to the challenges of the future. Some previous works discussing AI evaluation 
\cite{newell1976computer,
gaschnig1983evaluation,
rothenberg1987evaluating,
geissman1988verification,
decker1989evaluating,
langley1987research,
langley2011changing,
buchanan1988artificial,
simon1995artificial,
baldwin1995process,
falkenauer1998method,
langford2005clever,
LeggHutterTests2007,
whiteson2011protecting,
drummond2010warning,
anderson2011robotics,
madhavan2009performance,
schlenoff2011performance
} are relatively old, non-comprehensive, restrictive to a specific area of AI, limited to one particular approach and/or focussed on the experimental methodology rather than what is being measured and how. Nonetheless, we will refer to many of these works along the text. 

Some ideas of the old analysis still hold today. For instance, in \cite{cohen1988evaluation} we find criteria for evaluating research problems, methods, implementations, experiments' design, and evaluation of the experiments. In the criteria for experiments' design, we see several of the topics we will address in the paper: ``1. How many examples can be demonstrated?'' (are they sufficient and qualitative different and illustrative?), ``2. Should the program's performance be compared to a standard?'', ``3. What are the criteria for good performance?'', ``4. Does the program purport to be general (domain-independent)?'' (do the domains being tested constitute a representative class?), and ``5. Is a series of related programs being evaluated?''. Other statements in \cite{cohen1988evaluation} are not so up-to-date, and show that there has been an improvement in AI evaluation. For instance, we found the recommendation ``that editors, program committees, and reviewers should begin to insist on evaluation.''. 
Today this recommendation has been generalised (e.g., \cite{conrad2013significance} report that more than 60\% of ICAIL papers in 1987 did not have any evaluation in front of 20\% in 2011). Hence, a lack of evaluation is no longer the problem. However, there is still a great deal of disaggregation, many ad-hoc procedures, bad habits and loopholes about what is being measured and how is being measured. In this paper, the focus will be set on these issues.


We will start with a state of the art of the task-oriented evaluation approach in AI, by far more common in AI research. The notion of performance is relatively easy to determine as it is directly linked to the set or class of problems we are interested in for the evaluation. Nonetheless, we will identify several problems, most of them derived from the confusion of a task definition with its evaluation. An appropriate sampling procedure from the class of problems defining the task is not always easy. We will give some hints to derive better evaluation protocols.  
With this perspective we will argue that white-box evaluation (by algorithm inspection) is becoming less predominant in AI, and we will focus the rest of the paper to black-box evaluation (by behaviour). We will distinguish three types of behavioural evaluation: by human discrimination (performing a comparison against or by humans), problem benchmarks (a repository or generator of problems) and by peer confrontation (1-vs-1 or multi-agent `matches'). We will survey some of the competitions and repositories in these three categories and highlight some problems in how these evaluation settings are held and used.

In a second part of the paper, we will pay attention to the more elusive and challenging problem of ability-based evaluation. 
The three types of evaluation seen for task-oriented evaluation are not directly applicable, as we now do not want to evaluate systems for what they do but for what they are able to (learn to) do. In other words, we are looking for signs or indications that show that the system has a certain ability. One idea that has been around since the inception of AI is to use human (or animal) intelligence tests, such as the IQ-tests used in psychometrics. Each particular test tries to identify a series of exercises that are representative (necessary and sufficient) for a given ability. We will briefly discuss their use and possible adaptation for the evaluation of AI systems. A quite different approach is based on algorithmic information theory (AIT), where problem classes and their difficulty are derived from computational principles. In this way, we are sure about what we are actually evaluating. Also, exercise generators can be derived from first principles. 

While task-oriented evaluation is opposed to ability-oriented evaluation in this paper, we can have a more gradual view in terms of task classes that go from specific to general. Also, we will analyse a more unified view that integrates the different evaluation paradigms and procedures that we find in many disciplines, depending on the subject that is being measured. This view, known as `universal psychometrics', is based on the notion of `universal test'. This unified approach makes it possible that the schemas that were identified for the task-based evaluation can be generalised to the ability-based evaluation problem.

The rest of the paper goes along the organisation described above, with two parts: section \ref{sec:task-oriented}, focussing on task-oriented evaluation and section \ref{sec:ability-oriented}, focussing on ability-oriented evaluation. This is followed by the conclusions, which feature some guidelines about how competitions and problem generators can be improved, integrated or overhauled for a more robust and efficient AI system evaluation.

\section{Task-oriented evaluation}\label{sec:task-oriented}


AI is a successful discipline. The range of applications has been greatly enlarged over the years. We have successful applications in computer vision, speech recognition, music analysis, machine translation, text summarisation, information retrieval, robotic navigation and interaction, automated vehicles, game playing, prediction, estimation, planning, automated deduction, expert systems, etc. (see, e.g., \cite{Russell-Norvig2009}). Most of these application problems are specific. This implies that the goals are clear and that researchers can focus on the problem. This does not mean that we are not allowed to use more general principles and techniques to solve many of these problems, but that the task is sufficiently specific so that systems can be specialised for these systems. For instance, robotic navigation of a Mars rover can share some of the techniques with a driverless car on Earth, but the final application is extremely specialised in both cases.

This specialisation leads to an application-specific (task-oriented) evaluation. 
In fact, going from an abstract problem to a specific task is encouraged: ``refine the topic to a task'', provided it is ``representative'' \cite{cohen1988evaluation}. 
Given a precise definition of the task, we only need to define a notion of performance from it. Clearly, we measure performance, and not intelligence. In fact, many of the most successful AI systems solve each problem in a way that is different to the way humans solve the same problem. Also, AI systems usually include a great amount of built-in programming and knowledge for the task. It is not unfair to say that we evaluate the researchers that have designed the system rather than the system itself. For instance, we can say that it was the research team after Deep Blue \cite{deepblue2002} (with the help of a powerful computer) who actually defeated Kasparov.

Disregarding who is to praise for each new successful application, AI systems that address specialised problems with a clear performance should be easy to evaluate. The reality is not that straightforward, mostly because there are many different (and usually ad-hoc) evaluation approaches. Let us have a perusal over them.

\subsection{Types of performance measurement in AI}\label{sec:types}

An application, as described above, can be characterised by a set of problems, tasks or exercises $M$. In order to evaluate each exercise $\mu \in M$ we can get a measurement $R(\pi, \mu)$ of the performance of system $\pi$. Measurements can be imperfect. Also, the system, the problem or the measurement may be non-deterministic. As a result, it is usual to work with the expected value of the performance of $\pi$ as $\mathbb{E}[R(\pi, \mu)]$.

The definition of $M$ and $R$ does not specify how we want to aggregate the results when $M$ has more than one problem. The most common approaches are\footnote{Worst-case performance and best-case performance are special cases of a rank-based aggregation (using the cumulative distribution of results), with other possibilities such as the median, the first decile, etc. Rank-based aggregation, especially worst-case performance, is more robust to systems getting good scores on many easy problems but doing poorly on the difficult problems.}

\begin{itemize}
\item Worst-case performance\footnote{Note that this formula does not have the size of the instance as a parameter, and hence it is not comparable to the usual view of worst-case analysis of algorithms.}: \[\Phi_{min}(\pi, M) = \min_{\mu \in M} \mathbb{E}[R(\pi, \mu)] \]
\item Best-case performance: \[\Phi_{max}(\pi, M) = \max_{\mu \in M} \mathbb{E}[R(\pi, \mu)] \]
\item Average-case performance: \begin{equation}\Phi(\pi, M, p) = \sum_{\mu \in M} p(\mu) \cdot \mathbb{E}[R(\pi, \mu)] \label{eq:average} \end{equation}
where $p$ is a probability distribution on $M$.
\end{itemize}
\noindent It is assumed that the magnitudes of $R$ for different $\pi \in M$ are commensurate. For instance, if $R$ can range between 0 and 1 for problem $\mu_1$ but ranges between 0 and 10,000 for problem $\mu_2$, the latter will have a much higher weight and will dominate the aggregation. This is not necessarily wrong, e.g., if they are measured with the same unit (e.g., euros). In general, however, $R$ is a construct that needs to be normalised. The choice of a performance metric that is sufficiently normalised such that the results are commensurate is not always easy, but possible to some extent (see, e.g., \cite{whiteson2011protecting}).

At this point, it is pertinent to make a comment about the well-known no-free-lunch (NFL) theorems 
\cite{wolpert-macready1995no,NFLlearning,wolpert2012no}, as these theorems are usually misunderstood. These theorems state that given all possible problems, {\em under some particular distributions}, no method can work better than any other on average. 
The argument to support this interpretation is that, considering all problems, if method $\pi_A$ is better than method $\pi_B$ for one problem then $\pi_B$ will be worse than $\pi_A$ for another problem. 
Some people have even interpreted that research in AI (including search and optimisation problems in computer science) is futile. 
However, the NFL theorems can only be applied when the assumptions hold. 
The conditions state that $M$ must be infinite and include all possible problems. Also, the problems can be shuffled without affecting the probability, which can be expressed as ``block uniformity'' \cite{igel2005no}, for which the uniform distribution would be a special case.  
Nonetheless, these conditions are not plausible if the problems are taken from the real world. It is unrealistic to assume that the problems we face are taken from a series of random bits, or that a problem, and its opposite problem (whatever it is) are equally probable. 
Many other distributions are much more plausible. A universal distribution \cite{solomonoff1964,Li-Vitanyi08}, e.g., which is consistent with the idea that problems are generated by physical laws, processes, living creatures, etc., states that random (incompressible) problems are less likely.
So, for many distributions $p$, the conditions of the NFL do not hold and we have that there can be methods $\pi_A$ and $\pi_B$ such that: $\Phi(\pi_A, M, p) > \Phi(\pi_B, M, p)$. In fact, there can be optimal methods for inductive inference \cite{lattimore2013no}, some free lunches for co-evolution \cite{wolpert2005coevolutionary}, and other areas, 
although it seems that for optimisation the free lunches are very small \cite{everitt2014free}.

After this clarification, it is relevant to determine how $R$ is going to be obtained. 
For relatively simple solutions, we can analyse the code or the algorithm of the system $\pi$. If the code can be well understood then its computational properties and behaviour can be clearly determined. We use the term `white-box' evaluation when $R$ is inferred through program inspection or algorithm analysis. White-box evaluation is powerful because we can obtain $R$ theoretically for a given agent $\pi$ and a problem class $M$ (provided both are defined theoretically).  
One common type of problems that are evaluated with a white-box approach are those where the solution to the problem has to be correct or optimal (i.e., perfect). In this case, the performance metric $R$ is defined in terms of time and/or space resources. This is the case of classical computational complexity theory. Worst-case analysis is more common than average-case analysis, although the latter has also become popular recently \cite{knuth1973sorting,levin1986average,goldreich2007special}.
Nonetheless, many AI problems are so challenging nowadays that perfect solutions are no longer considered as a constraint. Instead, approximate solvers are designed to optimise a performance metric that is defined in terms of the level of {\em error} of the solution and the time and/or space resources. In this case, the use of an average-case analysis is more common, although worst-case analysis can also be studied under some paradigms (e.g., Probably Approximately Correct learning \cite{valiant1984theory}). 
In agent theory, the behaviour of the agent (and its properties) can be analysed under some paradigms such as Belief-Desire-Intention (BDI) agents (see, e.g., a testability approach in \cite{BDItest2014}). 
The theoretical analysis of `white-box' evaluation has also been applied to games. For instance, in board games, algorithms can be derived and analysed whether they are optimal, such as noughts and crosses (tic-tac-toe) and English draughts (checkers), the latter solved by Jonathan Schaeffer \cite{schaeffer2007checkers}. 
Finally, in game theory, the expected pay-off plays the role of $R$ and optimal strategies can be determined for some simple games, as well as equilibria and other properties. In games, some results can be obtained independently of the opponent, but others are only true if we also know the algorithm that the other players are using (so it becomes a double `white-box' approach to evaluation).

As AI systems become more sophisticated, white-box assessment becomes more difficult, if not impossible, because the unpredictability of complex systems. Many AI systems incorporate many different techniques and have stochastic behaviours. This is also in agreement with a view of AI as an experimental science \cite{buchanan1988artificial,simon1995artificial}. As a result, a black-box approach, is taken\footnote{The distinction between white and black box can be enriched to consider those problems where the solution must be accompanied by a verification, proof or explanation \cite{HernandezOrallo00b,alpcan2014}.}. This means that $R$ is obtained exclusively from the behaviour of the system in an empirical way. In this case, average-case evaluation is usual\footnote{Although it is not uncommon, as we will see, that the set of problems from $M$ are chosen by the research team that is evaluating its own method, so the probability to choose from $M$ can be biased in such a way that it is actually a best-case evaluation.}.

There are many kinds of black-box assessment in AI, but we can group them into three main categories:

\begin{itemize}
\item Human discrimination: The assessment is made by and/or against humans through observation, scrutiny and/or interview. Although it can be based on a questionnaire or a procedure, the assessment is usually informal and subjective. 
This type of evaluation is common in psychology, ethology and comparative psychology. In AI, this kind of evaluation is not very usual, except for the Turing Test and variants, as we will discuss later on.
\item Problem benchmarks: The assessment is performed against a collection or repository of problems ($M$). This approach is very frequent in AI, where we have problem libraries, repositories, corpora, etc. It is also usual in psychology and comparative psychology, although in these areas the tests are not publicly available to the systems that are to be evaluated. This has been proposed or suggested in AI occasionally (e.g., the ``secret generalized methodology'' \cite{whiteson2011protecting}). For instance, $M$ can be generated in real time using a problem generator, which actually defines $M$ and $p$. 
\item Peer confrontation: The assessment for (multi-agent) games is performed through a series of (1-vs-1 or $n$-vs-$n$) matches. The result is relative to the other participants. Given this relative value, in order to allow for a numerical comparison, sophisticated performance metrics can be derived (e.g., the Elo system in chess \cite{elo1978rating}).
\end{itemize}
\noindent The combination of some of the above is also common for evaluation. In what follows, we analyse each of the three categories in more detail.

\subsection{Evaluation by human discrimination}

In this first category we include the evaluation approaches that are performed by a comparison with or by humans. The Turing Test \cite{turing1950,oppydowe2011} is a case in which there is both comparison against humans and evaluation by human judges. While the `imitation game' was introduced by Turing as a philosophical instrument in his response to nine objections against machine intelligence, the game has been (mis-)understood as an actual test ever since, with the standard interpretation of one human, one machine pretending to be a human, and a human interrogator through a teletype acting as a judge. The latter must tell which one is the machine and the human.

Not only has the game been taken as an actual test, but it has had several implementations, such as the Loebner Prize\footnote{\url{http://www.loebner.net/Prizef/loebner-prize.html}.}, held every year since 1991. Despite the criticisms of how this prize is conducted and its interpretation through the years, there have been more implementations. In 2014, Kevin Warwick, professor at the University of Reading organised a similar competition that took place at the Royal Society in London. Even if the results were not significantly different to previous results of the Loebner Prize (or even what Weizenbaum's ELIZA was able to do fifty years ago \cite{Weizenbaum66}), the over-reaction and publicity of this outcome were preposterous. The reputation of the implementations of the Turing Test was (further) stained with statements such as this: ``If a computer is mistaken for a human more than 30\% of the time during a series of five minute keyboard conversations it passes the test. No computer has ever achieved this, until now. Eugene managed to convince 33\% of the human judges (30 judges took part  [...]) that it was human.'' \cite{pressrelease2014}. And Warwick goes on: ``We are therefore proud to declare that Alan Turing's Test was passed for the first time. [...] This milestone will go down in history as one of the most exciting''.

Is the imitation game a valid test? Even assuming that the times and thresholds are stricter than the previous incarnations, the Turing Test has many problems as an intelligence test. First, it is a test of humanity, relative to human characteristics (i.e., anthropocentric).
It is neither gradual nor factorial and needs human intervention (it cannot be automated). If done properly, it may take too much time. Even so, as we have seen, it can be gamed by non-intelligent chatterbox. As a result, the Turing Test is neither a sufficient nor a necessary condition for intelligence. 
Despite the criticism, the Turing Test still has many advocates \cite{proudfoot2011anthropomorphism}. It is also an inspiration for countless philosophical debates and has led to connections with other concepts in AI or computation \cite{manchester2012,AISB-AICAP2012b}.

In any case, Turing is not to be blamed by a failure of the Turing Test as a useful test to evaluate AI systems. Turing did not conceive the test as a practical test to measure intelligence up to and beyond human intelligence. He is not to blame for a philosophical construct that has had a great impact in the philosophy and understanding of machine intelligence, but originally a negative impact on its measurement.

Does this mean that we should discard the idea of evaluating AI systems by human judges or by comparing with humans? Not at all. Recently, there have been variants of the Turing Test (Total Turing Tests \cite{schweizer1998truly}, Visual Turing Tests including sensory information, Toddler Turing Tests \cite{alvarado2002beyond}, robotic interfaces, virtual worlds, etc. \cite{mueller2008adapting,hingston2010new})  that may be useful for chatterbot evaluation, personal assistants and videogames. It is within the area of videogames where the notion of `believability' has appeared, which is understood as the property of a bot of looking `believable' as a human \cite{livingstone2006turing,hingston2012believable}. This term is interesting, as it clearly detaches these tests from the evaluation of intelligence. In videogames, there are applications where we want bots that can fool opponents into thinking that they are just another human player. Other highly subjective properties may also be of interest: enjoyability, resilience, aggressiveness, fun, etc.

Finally, there is a kind of test that is related to the Turing Test, the so-called CAPTCHA (Completely Automated Public Turing test to tell Computers and Humans Apart) \cite{von2004telling,von2008recaptcha}.
It is said to be a `reverse Turing Test' because the goal is to tell computers and humans apart in order to ensure that an action or access is only performed by a human (e.g., making a post, registering in a service, etc.). CAPTCHAs are quick and practical, omnipresent nowadays. However, they are designed according to the tasks that are solved by the current state of AI technology. At present, for instance, a common CAPTCHA is a series of distorted letters, which are usually easy to recognise by humans but not by machines (e.g., current OCR systems struggle). Logically, when character recognition systems and other techniques improve, current CAPTCHAs are broken (see, e.g.,  \cite{bursztein2014end}), and CAPTCHAs need to be updated to more distorted words or to other tasks that are beyond AI technology. 
Similarly, the detection of bots in social networks (sybils) and crowdsourcing platforms rely on tests that are variants of CAPTCHAs, the Turing Test, or the observation and analysis of user profiles and behaviour \cite{chu2010tweeting,chu2010tweeting,wang2012social}.

\begin{table}
\begin{center}
\begin{tabular}{ll}
\hline
Evaluation Setting    & Description \\ \hline
Loebner Prize\tablefootnote{\url{http://www.loebner.net/Prizef/loebner-prize.html}} & General Turing Test implementation. \\
U. of Reading TT 2014\tablefootnote{\url{http://www.reading.ac.uk/news-and-events/releases/PR583836.aspx}} & General Turing Test implementation. \\
BotPrize\tablefootnote{\url{http://botprize.org/}} & Contest about bot believability in videogames. \cite{livingstone2006turing,hingston2012believable} \\
Robo Chat Challenge\tablefootnote{\url{http://www.robochatchallenge.com/}} & Chattering bots competition. \\
CAPTCHAs\tablefootnote{\url{http://www.captcha.net/}} & Spotting bots in applications requiring humans. \cite{von2004telling,von2008recaptcha} \\
Humies awards\tablefootnote{\url{http://www.human-competitive.org}} & Human-competitive results using genetic and evolutionary computation. \cite{koza2010human} \\
Graphics Turing Test & Tell between a computer-generated virtual world and a real camera. \cite{mcguigan2006graphics,borg2012practical} \\\hline
\end{tabular}
\caption{List of some evaluation settings in the human-discrimination category.}\label{tab:human-discrimination}
\end{center}
\end{table}

Table~\ref{tab:human-discrimination} includes a selection of evaluation settings under the human-discrimination category. 
As it is not possible to go into the details of all of them because of brevity, let us choose one that is most representative and with a strong future projection: the BotPrize competition, which has been held since 2008. This contest awards the bot that is deemed as more believable (playing like a human) by the other (human) players. The competition uses a first-person shooter videogame, the DeathMatch game type, as used in Unreal Tournament 2004. It is important to clarify that the bots do not process the image but receive a description of it through textual messages in a specific language through the GameBots2004 interface (Pogamut). For the competition, chatting is disabled (as it is not a chatbot competition). There is a ``judging gun'' and the human judges also play, trying to play normally (a prize for the judges exists for those that are considered more ``human'' by other judges).

Some questions have been raised about how well the competition evaluates the believability of the participants. For instance, believability is said to be better assessed from a third-person perspective (judging recorded video of other players without playing) than with a first-person perspective \cite{togelius2012assessing}.
The reason is that third-person human judges can concentrate on judging and not on not being killed or aiming at high scores. 
Actually, this third-person perspective is included in the 2014 competition using a crowdsourcing platform \cite{llargues2014artificial} so that the 2014 edition incorporates the two judging systems: the First-Person Assessment (FPA), using the BotPrize in-game judging system, and the Third-Person Assessment (TPA), using a crowdsourcing platform. Another issue that could be considered in the future is a richer (and more challenging) representation of the environment, closer to the way humans perceive the images of the game (such as the graphical processing required for the Arcade Learning Environment \cite{bellemare13arcade} or the General Video Game Competition \cite{Schaul2014}).

Finally, as a summary of the limitations and potentials of the human-discrimination category, we first acknowledge that some variants are being useful. However, the format differs significantly from a standard Turing Test. For instance, the human-discrimination approach to evaluation can be just solved by a more traditional interview format with a procedure or storyline (as in psychology interviews), or by an evaluation through observation (using a committee of dedicated judges). This casts doubts about whether evaluation by imitation using the standard interpretation of the Turing Test is practical for task-oriented evaluation in AI. It is the concept that is useful, and deserves being adapted to several applications.

\subsection{Evaluation through problem benchmarks}\label{sec:benchmarks}

In this very common approach to evaluation, $M$ is defined as a set of problems. This fits equation~\ref{eq:average} perfectly. Necessarily, the quality of the evaluations depends on $M$ and how exhaustively this set is explored. There are other issues that could compromise the quality of the measurement. For instance, when $M$ is a {\em public} problem repository and is not very large, we have that the systems can specialise for $M$. Also, the solutions may also be available beforehand, or  can be inferred by humans, so the systems can embed part of the solutions. In fact, a system can succeed in a benchmark with a small size of $M$ by using a technique known as the ``big switch'', i.e., the system recognises which problem is facing and uses the hardwired solution for that specific exercise. Things can become worse if the selection of examples from $M$ is made by the researchers themselves (e.g., the usual procedure in machine learning of selecting 10 or 20 datasets from the UCI repository \cite{uci}, as we will discuss below). In general, the size of $M$ and a bona fide attitude to research somewhat limit these concerns. Nonetheless, it is generally acknowledged that most systems actually embed what the researchers have learnt from $M$. In a way, these benchmarks actually evaluate the researchers, not their systems.

The above-mentioned problem is known as `evaluation overfitting' \cite{whiteson2011protecting}, `method overfitting' problem \cite{falkenauer1998method} or ``clever methods of overfitting'' \cite{langford2005clever}. To avoid or reduce this problem, it is much better if $M$ is very large or infinite, or at least the problems are not disclosed until evaluation time. Problem generators are an alternative. However, it is not always easy to generate a large $M$ of realistic problems. Generators can be based on the use of some prototypes with parameter variations or distortions. These prototypes can be ``based on reality'', so that the generator ``takes as input a real domain, analyses it automatically and generates deformations [...] that follow certain high-level characteristics'' \cite{drummond2010warning}. 
More powerful and diverse generators can be defined by the use of problem representation languages. 
A general and elegant approach is to determine a probabilistic or stochastic generator (e.g., a grammar) of problems, which directly defines the probability $p$ for the average-case performance equation~\ref{eq:average}.
Nonetheless, it is not easy to make a generator that can rule out unusable or Frankenstein problems. In other domains, problems are taken from real life (e.g., pedestrian detection), and having a large number of labelled examples is very expensive. Virtual simulators are becoming common to create problems \cite{vazquez2013virtual}.

When the set of problems is large or generated, we clearly cannot evaluate AI systems efficiently with the whole set $M$. So we need to do some sampling of $M$. It is at this point when we need to distinguish the benchmark or problem definition from an effective evaluation. Assume we have a limited number of exercises $n$ that we can administer. The goal will be to reduce the variance of the measurement given $n$. One naive approach is to sort $M$ by decreasing $p$ and evaluate the system with the first $n$ exercises. This maximises the accumulated mass for $p$ for a given $n$. One problem about this procedure is that it is highly predictable. Systems will surely specialise on the first $n$ exercises. Also, this approach is not very meaningful when $R$ is non-deterministic and/or not completely reliable. Repeated testing may be necessary, which raises the question of whether to explore a higher $n$ or to perform more repetitions. 

\begin{figure}
\centering
\vspace{-1cm}
\includegraphics[width=0.48\textwidth]{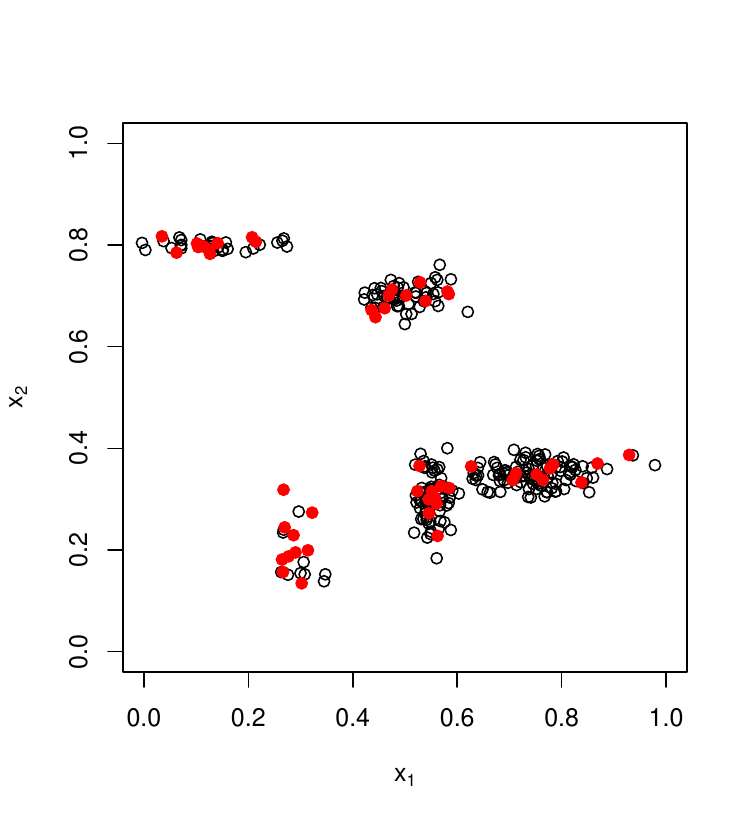} \hfill
\includegraphics[width=0.48\textwidth]{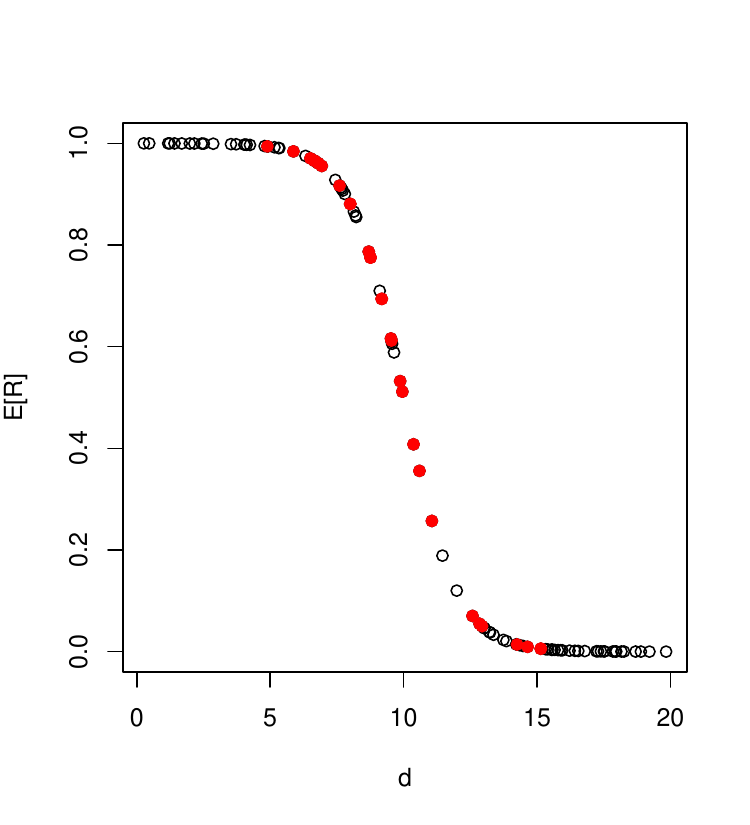} 
\vspace{-0.5cm}
\caption{Left: a repository $M$ with $|M|=300$ exercises shown with empty black circles. Two features $x_1$ and $x_2$ are used to describe the most relevant characteristics of the exercises (according to diversity). These features are used to cluster them into five groups. Next, cluster sampling is performed with a sample size of $n=50$. Clusters are of different size (60, 20, 70, 110, 30) but 10 samples (shown in solid red circles) are taken from each cluster. Because of the constant number of examples per cluster, in order to estimate $\Phi$, measurements for under-represented clusters are multiplied by their size. Right: a repository of $|M|=100$ exercises. A measure of difficulty $d$ has been derived that is monotonically decreasing with (estimated) expected performance (for a group of agents or for the problem overall). Only $n=30$ exercises are sampled in the area where the results may be most informative.}
\label{fig:sampling}
\vspace{-0.3cm}
\end{figure}

Random sampling using $p$ seems to be a more reasonable alternative. As said above, if $R$ is non-deterministic and/or subject to measurement error, then random sampling can be with replacement. If $M$ and $p$ define the benchmark, is probability-proportional sampling on $p$ the best way to evaluate systems? The answer is no, in general. 
There are better ways of approximating equation~\ref{eq:average}. The idea is to sample in such a way that the diversity of the selection is increased. This `diversity-driven sampling'' is related to several kinds of sampling, such as importance sampling \cite{srinivasan2002importance}, 
 stratified sampling \cite{cochran2007sampling} 
 and other forced Monte Carlo procedures. The key issue is that we use a {\em different} probability distribution for sampling. 
Although there are many ways of obtaining a `diverse' sample, we just highlight two main approaches that can be useful for AI evaluation:

\begin{itemize}
\item Information-driven sampling: Assume that we have a similarity function $sim(\mu_1, \mu_2)$, which indicates how similar (or correlated) exercises $\mu_1$ and $\mu_2$ in $M$ are. 
In this case, we need to sample on $M$ such that the accumulated mass on $p$ is high and that diversity is also high. 
The rationale is that if $\mu_1$ and $\mu_2$ are very similar, using one of them can `fill the gap' of the other, and we can assume as if both $\mu_1$ and $\mu_2$ had been explored, actually accumulating $p(\mu_1) + p(\mu_2)$. One possible way of doing this is 
by cluster sampling. 
Information-driven sampling suffers from the need of defining the similarity function $sim$. An alternative is to derive $m$ features that describe the exercises, so creating an $m$-dimensional space where distances and other topological information can be used to support the notion of diversity (and performing clustering). An example of this procedure is shown in Figure \ref{fig:sampling} (left). 
\item Difficulty-driven sampling. A set $M$ can contain very easy and very challenging problems. Using easy problems for good systems or difficult problems for bad systems is not very optimal. 
The idea to optimise the evaluation is to choose a range of difficulties for which the evaluation results may be informative (or to give higher probability to exercises inside this range), as in Figure \ref{fig:sampling} (right). This procedure is done to a greater or lesser degree in many evaluations and benchmarks in AI. In fact, more challenging problems are usually added over the years, as the systems are able to solve the easy problems (that soon become `toy problems'). One of the crucial points of difficulty-driven sampling is the definition of a difficulty function  $d: M \rightarrow \mathbb{R}^+$. 
Ideally, we would like that for every $\pi, \Phi(\pi, \mu_1, p) > \Phi(\pi, \mu_2, p)$ iff $d(\mu_1) < d(\mu_2)$. 
In practice, this condition is too strong, and more flexible characterisations are expected, such as that for every $\pi$, and two difficulties $a$ and $b$ such that $a \leq b$ we have that $\Phi(\pi, M_a, p) \geq \Phi(\pi, M_b, p)$  (where $M_a$ denotes all the exercises in $M$ of difficulty $a$). This could still too strong and we may use a relaxed version such that 
for every $\pi$, there is a $t$ such that for all $a$ and $b \geq a+t$: $\Phi(\pi, M_a, p) \geq \Phi(\pi, M_b, p)$. 
In experimental sciences, we have a population-based view of difficulty such that $d(\mu)$ is monotonically decreasing on $\mathbb{E}_{\pi \in \Omega} [\Phi(\pi, \mu, p)]$, where $\Omega$ is a population of subjects, agents or systems that are evaluated for the same problem class. In fact, Item Response Theory \cite{embretson2000item} in psychometrics follows this approach. Finally, we can derive the difficulty of a problem as a function of the complexity of the problem itself. The complexity metric can be specific to the application (such as the complexity for mazes in \cite{bagnall2005classification,Zatuchna-Bagnall09} or grid-world domains in \cite{sturtevant2012benchmarks}) or it can be a more general approach (e.g., Kolmogorov complexity). Note that some of the definitions of difficulty above would not be possible for a set $M$ and distribution $p$ if the conditions of the NFL theorem held. 
\end{itemize}

\noindent Both the information-driven sampling and the difficulty-driven sampling can be made adaptive. The first is represented by what is known by adaptive cluster sampling \cite{seber2013adaptive}  
and it is common in population surveys and many experimental sciences. However, when evaluating performance, it is difficulty-driven sampling that has been used more systematically in the past, especially in psychometrics. In psychometrics, difficulty is inferred from a population of subjects (in the case of AI, this could be a set of solvers or algorithms). Instead of difficulty, items are analysed by proficiency, represented by $\theta$, a corresponding concept to difficulty from the point of view of the solver (higher problem difficulty requires higher agent proficiency).

\begin{figure}
	\centering
		\vspace{-1.2cm}
		\hspace{-0.5cm} 
		\includegraphics[width=0.50\textwidth]{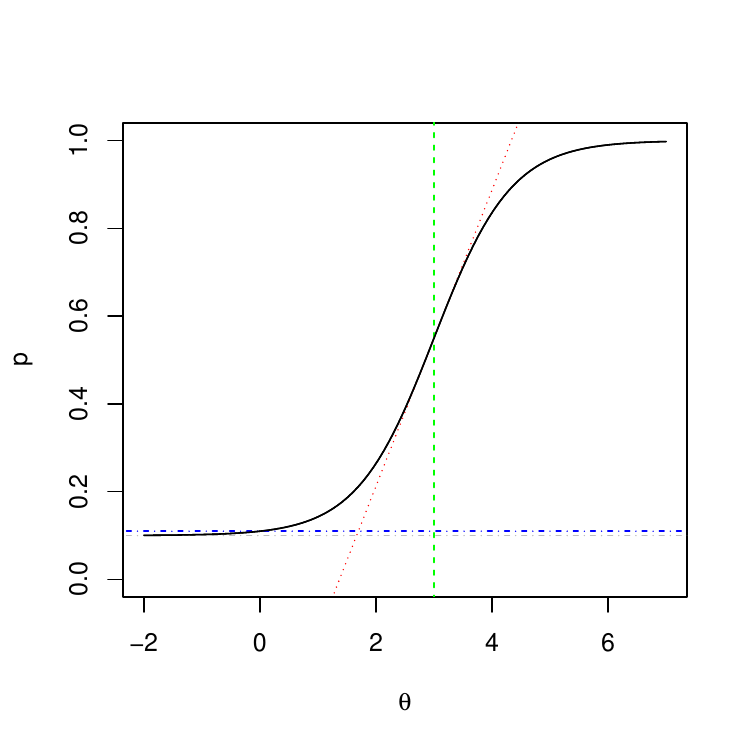} \hspace{-0.3cm}
		\includegraphics[width=0.50\textwidth]{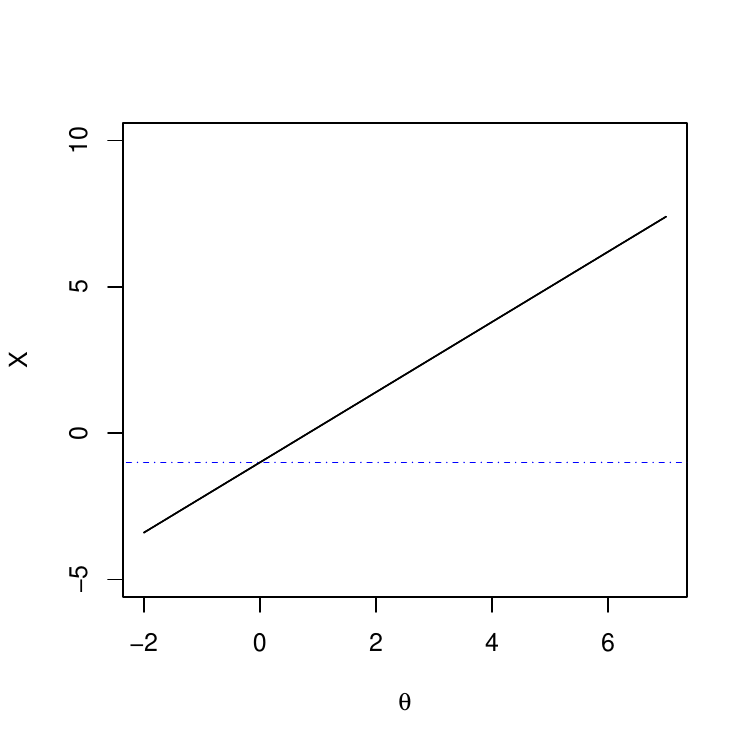} \hspace{-0.7cm}
		\vspace{-0.3cm}
	\caption{Left: item response function (or curve) for a binary score item with the following parameters for the logistic model: discrimination $a=1.5$, item location $b=3$, and chance $c=0.1$. The discrimination is shown by the slope of the curve at the midpoint: $a(1-c)/4$ (in dotted red), the location is given by $b$ (in dashed green) and the chance is given by the horizontal line at $c$ (in dashed-dotted grey), which is very close to the zero-proficiency expected result $p(\theta)=z$ (here at 0.11). Right: A linear model for a continuous score item with parameter $z=-1$ and $\lambda=1.2$. The dashed-dotted line shows the zero-ability expected result.}
	\label{fig:irc}
\end{figure}

Item response theory (IRT) \cite{embretson2000item} 
estimates mathematical models to infer the associated probability and informativeness estimations for each item. When $R$ is discrete or bounded, one very common model is the three-parameter logistic model, where the item response function (or curve) corresponds to the probability that an agent with proficiency $\theta$ gives a correct response to an item. This model is characterised as follows:
\[p(\theta) \triangleq c + \frac{1-c}{1+e^{-a(\theta-b)}} \]
\noindent where $a$ is the {\em discrimination} (the maximum slope of the curve), $b$ is the {\em difficulty} or item location (the value of $\theta$ leading to a probability half-way between $c$ and 1, i.e., $(1+c)/2$), and $c$ is the chance or asymptotic minimum (the value that is obtained by {\em random} guess, as in multiple choice items). 
The zero-ability expected result is given when $\theta=0$, which is exactly $z = c + \frac{1-c}{1+e^{ab}}$. 
Figure \ref{fig:irc} (left) shows an example of a logistic item response curve. 

For continuous $R$, if they are bounded, the logistic model above may be appropriate. On other occasions, especially if $R$ is unbounded, a linear model may be preferred \cite{mellenbergh1994,ferrando2009difficulty}:
\[X(\theta) \triangleq z + \lambda \theta + \epsilon \]
\noindent where $z$ is the intercept (zero-ability expected result), $\lambda$ is the loading or slope, and $\epsilon$ is the measurement error. Again, the slope $\lambda$ is positively related to most measures of discriminating power \cite{ferrando2012discriminating}.
Figure \ref{fig:irc} (right) shows an example of a linear item response curve.

Working with item response models is very useful for the design of tests, because if we have a collection of items, we can choose the most suited one for the subject (or population) we want to evaluate. According to the results that the subject has obtained on previous items, we may choose more difficult items if the subject has succeeded on the easy ones,  we may look for those items that are most discriminating (i.e., most informative) in the area we have doubts, etc. Note that discrimination is not a global issue: a curve may have a very high slope at a given point, so it is highly discriminating in this area, but the curve will almost be flat when we are far from this point. Conversely, if we have a low slope, then the item covers a wide range of difficulties but the result of the item will not be so informative as for a higher slope.

\begin{figure}
\centering
\vspace{-1cm}
\includegraphics[width=0.58\textwidth]{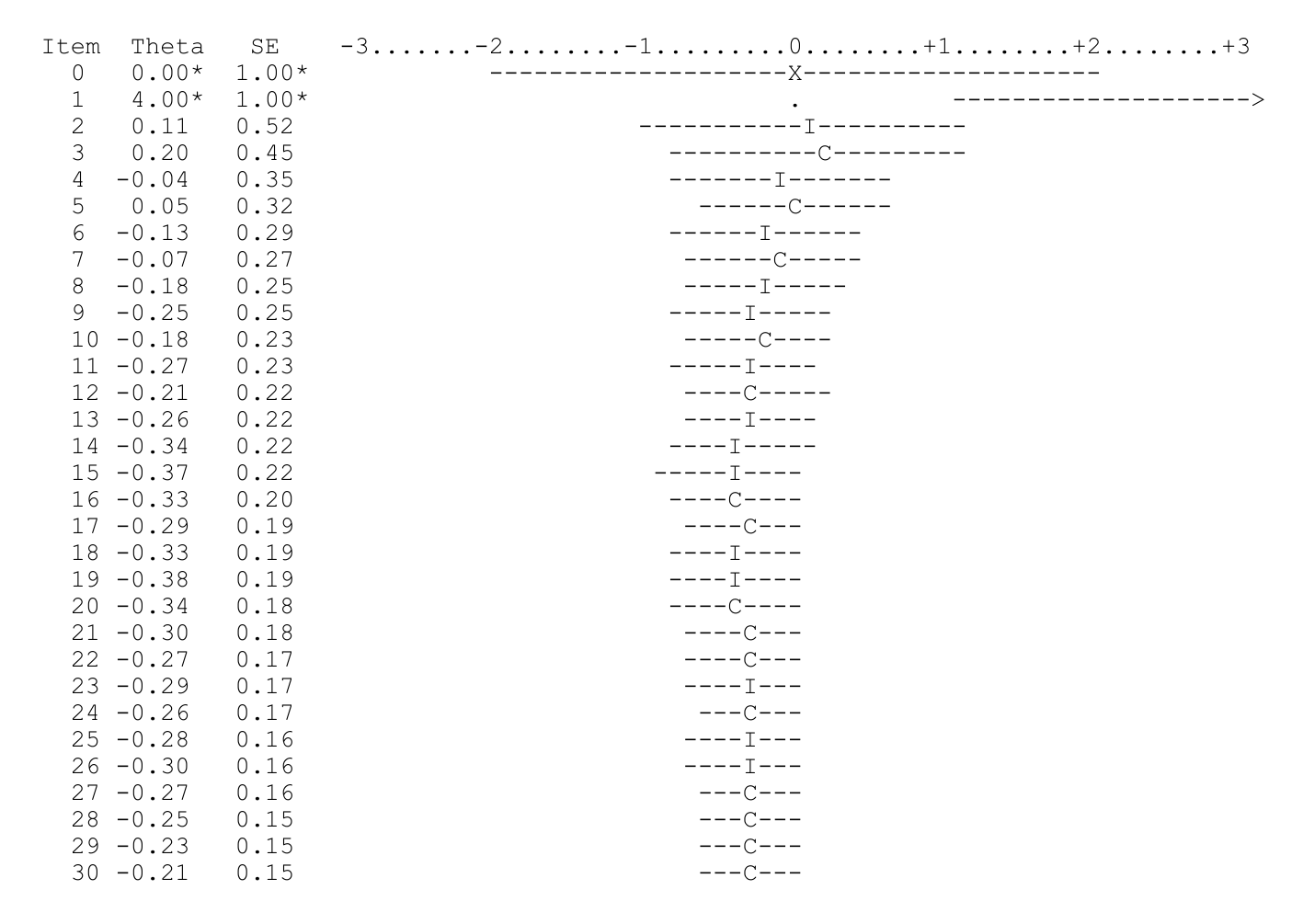} \hfill
\includegraphics[width=0.41\textwidth]{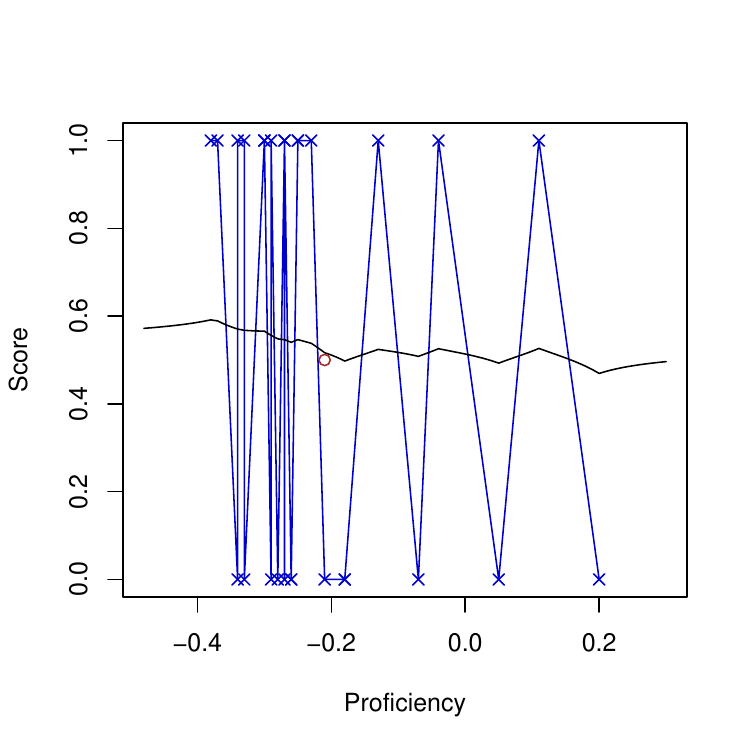} \hfill
\vspace{-0.6cm}
\caption{An example of an IRT-based adaptive test (freely adapted from 
\cite[Fig. 8]{weiss2011better}). Left: the process and proficiencies ({\em thetas}) used until convergence. The final proficiency calculated by the test was $-0.21$ with a standard error of $0.15$. 
Right: The results shown on a plot. The black curve shows a Euclidean kernel smoothing with a constant of 0.1.  }
\label{fig:cat}
\vspace{-0.3cm}
\end{figure}

Figure \ref{fig:cat} shows an example of an adaptive test using IRT. The sequence of exercise difficulties is shown on the left. The plot on the right shows that averaging the results (especially here, as the outcome of $R$ is discrete, either 0 or 1) makes the estimation of $\Phi$ more difficult with a non-adaptive test.

\begin{table}
\begin{center}
\begin{tabular}{ll}
\hline
Evaluation Setting    & Description \\ \hline
CADE ATP System Competition\tablefootnote{\url{http://www.cs.miami.edu/~tptp/CASC/}} & Theorem proving \cite{SS06-SoCASC} using the TPTP library \cite{SS98}. \\
Termination Competition\tablefootnote{\url{http://termination-portal.org/wiki/Termination_Competition_2014}} & Termination of term rewriting and programs \cite{marche2007termination}. \\
The reinforcement learning competition\tablefootnote{\url{http://www.rl-competition.org/}} & Reinforcement learning \cite{whiteson2010reinforcement}. \\
Syntax-guided synthesis competition\tablefootnote{\url{http://www.sygus.org/}} & Program synthesis \cite{alur2013syntax}. \\
International Aerial Robotics Competition\tablefootnote{\url{http://www.aerialroboticscompetition.org/}} & Pilotless aircraft competition. \\
DARPA Grand Challenge\tablefootnote{\url{http://archive.darpa.mil/grandchallenge04/index.htm}} & Autonomous ground vehicles.  \\
DARPA Urban Challenge\tablefootnote{\url{http://archive.darpa.mil/grandchallenge/}} & Driverless vehicles.  \\
DARPA Cyber Grand Challenge\tablefootnote{\url{http://www.darpa.mil/cybergrandchallenge/}} & Computer security. \\
DARPA Save the day\tablefootnote{\url{http://www.theroboticschallenge.org/}} & Rescue Robotic challenge \cite{jacoff2003test}. \\
The planning competition\tablefootnote{\url{http://ipc.icaps-conference.org/}} & Planning \cite{Planningcompetition}. \\
UCI\tablefootnote{\url{http://archive.ics.uci.edu/ml/}} and KEEL\tablefootnote{\url{http://sci2s.ugr.es/keel/datasets.php}} & Machine learning dataset repositories \cite{uci} \cite{alcala2010keel}. \\
PRTools\tablefootnote{\url{http://prtools.org/}} & Pattern recognition problem repository. \\
KDD-cup challenges\tablefootnote{\url{http://www.sigkdd.org/kddcup/index.php}} and kaggle\tablefootnote{\url{http://www.kaggle.com/}} & Machine learning and data mining competitions. \\
Plagiarism detection\tablefootnote{\url{http://pan.webis.de/}} & Plagiarism detection, authorship and social software misuse \cite{plagiariasm2013}. \\
The General Video Game Competition\tablefootnote{\url{http://www.gvgai.net/}} & General video game players \cite{Schaul2014}. \\
Hutter Prize\tablefootnote{\url{http://prize.hutter1.net/}} and related benchmarks\tablefootnote{\url{http://mattmahoney.net/dc/text.html
}} & Text compression. \\
Pedestrian benchmarks & Pedestrian detection \cite{geronimo2014datasets} \\
Europarl\tablefootnote{\url{http://www.statmt.org/europarl/}}, SE times corpus\tablefootnote{\url{http://www.statmt.org/setimes/}}, the euromatrix\tablefootnote{\url{http://matrix.statmt.org/matrix/info}} & Machine translation corpora \cite{starkie2006tenjinno}.  \\
Linguistic data consortium corpora\tablefootnote{\url{https://www.ldc.upenn.edu/new-corpora}} & NLP corpora. \\
The Arcade Learning Environment\tablefootnote{\url{http://www.arcadelearningenvironment.org/}} & Atari 2600 videogames (reinforcement learning) \cite{bellemare13arcade}. \\   
GP benchmarks\tablefootnote{\url{http://gpbenchmarks.org/}} & Genetic programming \cite{gp-benchmarks-2012,gp-benchmarks-2013}. \\
Pathfinding benchmarks\tablefootnote{\url{http://www.movingai.com/benchmarks/}} & Gridworld domains (mazes) \cite{sturtevant2012benchmarks}. \\
FIRA HuroCup\tablefootnote{\url{http://www.fira.net/contents/sub03/sub03_1.asp}} & Humanoid robot competitions \cite{anderson2011robotics}
 \\\hline
\end{tabular}
\caption{List of some evaluation settings in the problem-benchmarks category.}\label{tab:problem-benchmarks}
\end{center}
\end{table}

Table~\ref{tab:problem-benchmarks} includes a selection of evaluation settings in the problem benchmarks category. We see the variety of repositories, challenges and competitions. As it is impossible to survey all of them in detail, we will focus on one of them, perhaps the most widespread repository in computer science, the UCI machine learning repository \cite{uci}. Most of the discussion below is applicable to other repositories and, to some extent, to competitions and challenges in machine learning. 

The UCI repository includes many supervised (classification and regression) and some unsupervised datasets. The repository is publicly available and is regularly used in machine learning research. The usage procedure, which is referred as ``The UCI test'' \cite{macia2014towards} or the ``de facto approach" \cite{drummond2010warning}\cite{japkowicz2011evaluating}, follows the general form of equation~\ref{eq:average} where $M$ is the repository, $p$ is the choice of datasets and $R$ is one particular performance metric (accuracy, AUC, Brier score, F-measure, MSE, etc. \cite{ferri2009experimental,hernandez2012unified}). With the chosen datasets, several algorithms (where one or more are usually introduced by the authors of the research work) can be evaluated by their performance on the datasets. The aggregation over several datasets according equation~\ref{eq:average}, however, is not very common in machine learning, as there is the general belief that averaging the results for several datasets is wrong, as results are not commensurate (see, e.g., \cite{demvsar2006statistical}). We already discussed this issue in section \ref{sec:types} and saw that there are ways to normalise the performance metric or use some utility measure instead (e.g., what are the costs, in euros, of false positives and false negatives for each dataset) such that they can be aggregated. Nonetheless, statistical tests are the predominant and encouraged approach to evaluation validation by the machine learning research community.

``The UCI test" can be seen as a bona-fide mix of the problem benchmark approach and the peer confrontation approach. Even if there is a repository ($M$), only a few problems are chosen, and can be cherry-picked ($p$ is changing and arbitrary). Also, as the researchers' algorithm has to be compared with other algorithms, a few competitors are chosen, which can also be cherry-picked, without much effort on fine-tuning their best parameters. Finally, as the results are analysed by statistical tests, cross-validation or other repetition approaches are used to reduce the variance of $R(\pi, \mu, p)$ so that we have fewer ``ties''. This procedure frequently leads to claims about new methods being better than the rest. Many of these claims are, apart from uninteresting, dubious, even for papers in good venues. Nonetheless, the UCI repository is not to blame for this procedure, but a methodology where statistical significance for a few datasets is more valued than a commensurate average aggregate performance on a large collection of datasets.

As a result, there have been suggestions of a better use of the UCI repository. These suggestions imply an improvement of the procedure but also of the repository itself. For instance,  
UCI+, ``a mindful UCI'' \cite{macia2014towards}, proposes the characterisation of the problems in the UCI repository by a set of complexity measures from \cite{ho2002complexity}. 
This characterisation can be used to make samples that are more diverse and representative. Also, they discuss the notion of a problem being `challenging', trying to infer a notion of `difficulty'. In the end, an artificial dataset generator is proposed to complement the original UCI dataset. It is a distortion-based generator (similar to Soares's UCI++ \cite{soares2009uci++}). Finally, \cite{macia2014towards} suggest ideas about sharing and arranging the results of previous evaluations so that each new algorithm can be compared immediately with many other algorithms using the same experimental setting. 
This idea of `experiment database' \cite{vanschoren2012experiment} has already been set up. Openml\footnote{\url{http://openml.org/}.} 
\cite{van2013openml,vanschoren2014openml}, is an open science platform that integrates machine learning data, software and results
An automated submission procedure, such as Kaggle, if performed for a wide range of problems at a time, could be a way of controlling some of the methodological problems of how the UCI repository is used.

Although some of these improvements are in the line of better sampling approaches (more representative and more effective), there are still many issues about the way these repositories are constructed and used. The complexity measures could be used to derive how representative a problem is with respect to the whole distribution in order to make a more adequate sampling procedure (e.g., a clustering sampling). Also, a pattern-based generator instead of a distortion-based generator could give more control of what is generated and its difficulty. This could be done with a stochastic generative grammar for different kinds of patterns, as is usually done with artificial datasets, using Gaussians or geometrical constructs. 
Finally, if results are aggregated according to equation~\ref{eq:average}, the experimental setting and the use of repetitions should be overhauled. For instance, by using 20 different problems with 10 repetitions using cross-validation (a very common setting in machine learning experiments) we have less information than by using 200 different problems with 1 repetition. Choosing the least informative procedure only makes sense because of the way results are fitted into the statistical tests and also because repetitions usually involve less effort than preparing a large number of datasets.

Overall, even if the UCI repository and machine learning are very particular, many of the benchmarks in Table~\ref{tab:problem-benchmarks} suffer from the same problems about how representative the problems are (if $M$ is small) or how representative the sample is (if $M$ is large). 
%
Other problems are the estimation of task difficulty and whether $M$ is able to discriminate between a set of AI systems. Also, none of the benchmarks in AI is adaptive.

\subsection{Evaluation by peer confrontation}

In the evaluation by peer confrontation, we evaluate a system by confronting it to another system. This usually means that a match is played between peers. This is usual for games (including game theory) and part of multi-agent research. The results of each match (possibly repeated with the same peer) may serve as an estimation of which of the two systems is best (and how much). Nonetheless, the main problem about this approach is that the results are relative to the opponents. This is natural in games, as people are said to be good or bad at chess, for instance, depending of whom they are compared with. 

Despite this relative character of the evaluation, we can still see the average performance according to equation~\ref{eq:average}. In order to do this, we must first identify the set of opponents $\Omega$. Then, the set of problems $M$ is enriched (or even substituted) by the parametrisation of each single game (e.g., chess) with different competitors from $\Omega$. 
In 1-vs-1 matches we have that $|M|=|\Omega|-1$ (if we do not consider a match between a system and itself). In other multi-agent situations where many agents play at the same time, $|M|$ can grow combinatorially on $|\Omega|$. 

Nonetheless, for AI research, our main concern is about robustness and standardisation of results. For instance, how can we compare results between two different competitions if opponents are different? If these competitions are performed year after year, how can we compare progress?
If there are common players, we can use rankings, such as the Elo ranking \cite{elo1978rating}, or more sophisticated rating systems \cite{smith2002rating,masum2003turing}, to see whether there are progress. In fact, it would be very informative for AI competitions based on peer confrontation to keep all participants from previous editions in subsequent editions. However, this comes with a drawback, as systems could specialise to the kind of opponents that are expected in a competition. If a high percentage of competitors are inherited from previous editions, specialisation to those old (and bad) systems could be common. It is insightful to think how many of these issues are addressed in sport competitions. For instance, some tournaments adapt their matches according to previous information (by round, by ranking, etc.). In fact, a league may be redundant (for the same reasons why the information-driven or difficulty-driven sampling are introduced) and other tournament arrangements are more effective with almost the same robustness and much fewer matches.

As an alternative, games and multi-agent environments could be evaluated against standardised opponents. However, how can we choose a set of standardised opponents?
If the opponents are known, the systems can be specialised to the opponents. For instance, in an English draughts (checkers) competition, we could have players being specialised to play against Chinook, the proven optimal player \cite{schaeffer2007checkers}. Again, this ends up again in the design of an opponent generator. This of course does not mean a random player (which is usually very bad), but players that can play well. One option is to use old systems where some parameters are changed. Alternatively, a more far-reading approach is to define an agent language and generate players (programs) with that language. As it is expected that this generation will not achieve very good players (otherwise we would be facing a very simple problem), a possible solution is to give more information and resources to these standardised opponents to make them more competitive (e.g., in some applications these opponents could have more sophisticated sensor mechanisms or some extra information about the match that regular players do not have).

\begin{figure}
	\centering
		\vspace{-1.0cm}
		\hspace{-0.3cm} 
\includegraphics[width=0.5\textwidth]{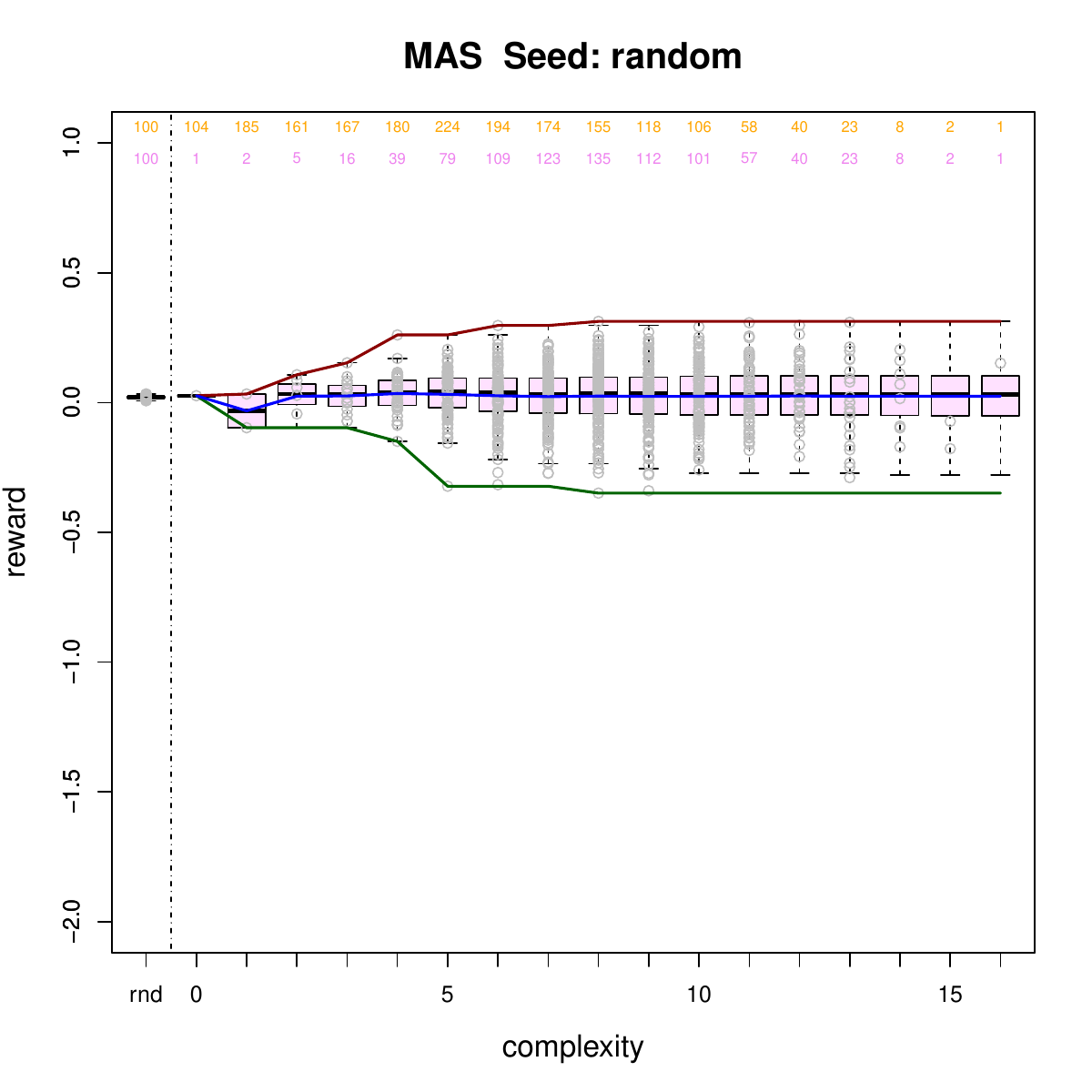}
		\includegraphics[width=0.5\textwidth]{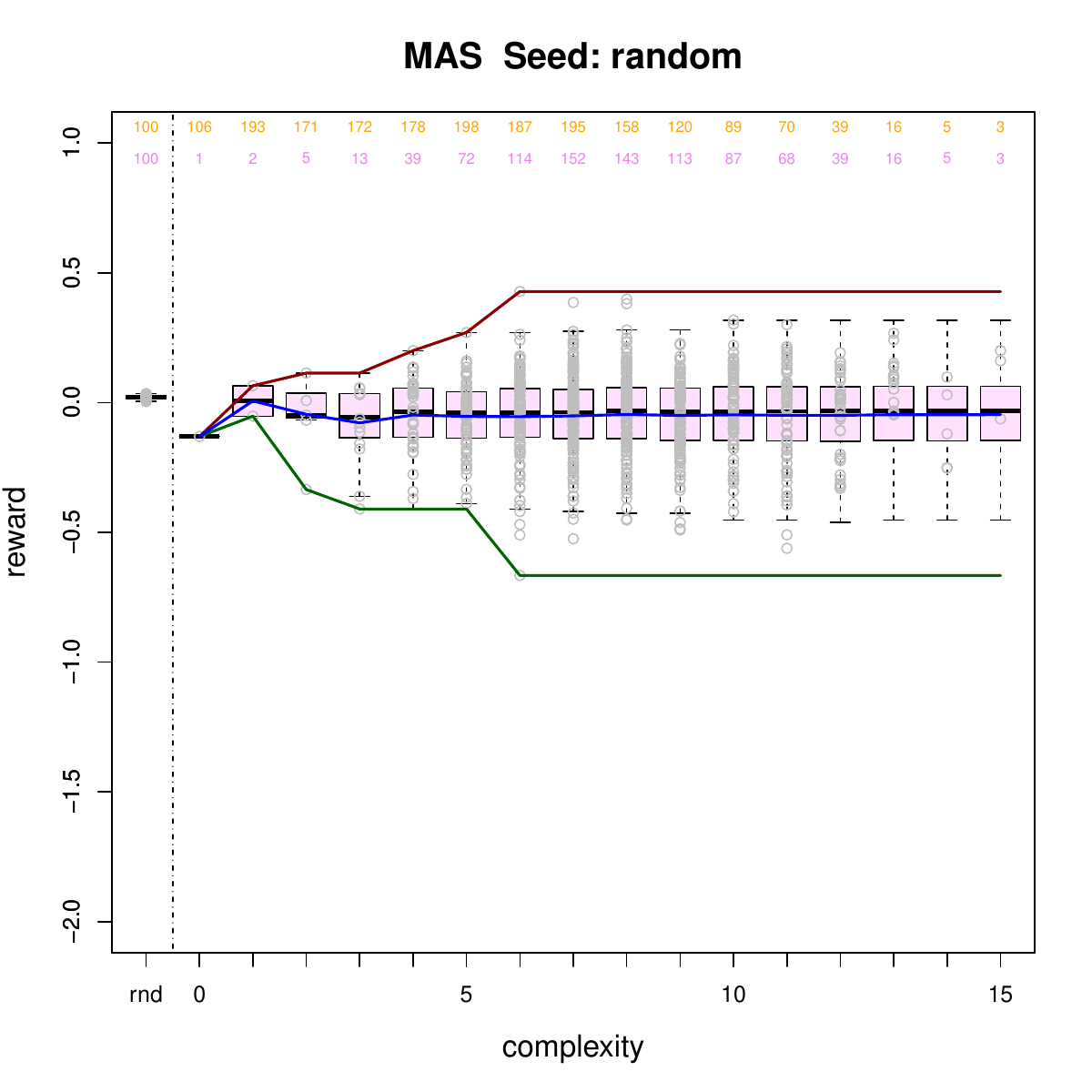}		                                       
\hspace{-0.4cm}
		\vspace{-0.2cm}
	{\caption{We show the distributions of reward (roughly corresponding to $R$ in this paper) for different configurations for the multi-agent system {\sffamily SCMAS} introduced in \cite{orallo2014JAAMAS}. Left: the plot shows the results when we confront each of the 2,000 policies with 50 different teams of competitors (with different seeds for the generator also). This means that we have 2,000 $\times$ 50 = 100,000 experiments (300 environment steps each). The results for a random agent (\sffamily{rnd}) are also shown for comparison. 
	Right: results when we choose the best 8 agents from the previous experiment. We see a wider range of results (but note that the average reward is lower).}\label{fig:MAS-distribution-single}}
\end{figure}

Be the set $\Omega$ composed of old opponents or generated opponents, we need to assess whether $\Omega$ is sufficiently challenging and whether it is able to discriminate the participants. For instance, some competitions in AI finally award a champion, but there is the feeling that the result is mostly arbitrary and caused by luck, as happens with many sport competitions\footnote{Statistical tests are not used to determine when a contestant can be said to be significantly better than another.}. How can we assess whether the set $\Omega$ has sufficiently difficulty and discriminating power? This is of course a hard problem, which has recently be analysed in \cite{orallo2014JAAMAS}. For instance, Figure \ref{fig:MAS-distribution-single} shows the distribution of results of an agent competing in a multi-agent system according to the complexity of the agent. The difficulty and discriminating power varies depending on the opponents (left vs right plots).

Table~\ref{tab:peer-confrontation} shows a sample of evaluation settings based on peer confrontation. Once again, because of obvious space constraints, we will just choose one representative and interesting case from the table. We will discuss the General Game Competition, which has been run yearly since 2005. According to the webpage\tablefootnote{\url{http://games.stanford.edu/}}, 
``general game players are systems able to accept descriptions of arbitrary games at runtime and able to use such descriptions to play those games effectively without human intervention. In other words, they do not know the rules until the games start''. Games are described in the language GDL (Game Description Language). The description of the game is given to the players. Different kinds of games are allowed, such as noughts and crosses (tic tac toe), chess, in static or dynamic worlds, with complete or partial information, with varying number of players, with simultaneous or alternating plays, etc. The competition consists of several rounds, qualifications, etc. 
For the competition, games are chosen ---non-randomly, i.e., manually by the organisers--- from the pool of games already described in GDL and new games are also newly introduced for the competition. As a result, game specialisation is difficult.

\begin{table}
\begin{center}
\begin{tabular}{ll}
\hline
Evaluation Setting    & Description \\ \hline
Robocup\tablefootnote{\url{http://www.robocup.org/}} and FIRA\tablefootnote{\url{http://www.fira.net}} & Robotics (robot football/soccer) \cite{kitano1997robocup,kim2004soccer}. \\
General game playing AAAI competition\tablefootnote{\url{http://games.stanford.edu/}} & General game playing using GDL \cite{genesereth2005general}. \\
World Computer Chess Championship\tablefootnote{\url{http://www.icga.org/}} & Chess. \\
Computer Olympiad\tablefootnote{\url{http://www.icga.org/}} & Board games. \\
Annual Computer Poker Competition\tablefootnote{\url{http://www.computerpokercompetition.org/}} & Poker. \\
Trading Agents Competition\tablefootnote{\url{http://tradingagents.eecs.umich.edu/}} & Trading agents \cite{Tradingagentcompetition,ketter2012competitive}. \\
Warlight AI Challenge\tablefootnote{\url{http://theaigames.com/competitions/warlight-ai-challenge/rules}} & Strategy games (Warlight). \\\hline
\end{tabular}
\caption{List of some evaluation settings in the peer-confrontation category.}\label{tab:peer-confrontation}
\end{center}
\end{table}

Despite being one of the most interesting AI competitions, there is still some margin for improvement. For instance, a more sophisticated analysis of how difficult and representative each problem is would be useful. 
For instance, several properties about the adequacy of an environment or game for peer-confrontation evaluation could be identified and analysed depending on the population of opponents that is being considered \arxiv{\cite{insa2014}}. 
Also, rankings (e.g., using the Elo system mentioned above) could be calculated, and former participants could be kept for the following competitions, so there are more participants (and more overlap between competitions). A more radical change would be to learn without the description of the game, as a reinforcement learning problem (where the system learns the rules from many matches). An adaptation between the general game playing and RL-glue, which is used in the reinforcement learning competition, to make this possible has been done in \cite{ayuso2012integration}. 

Summing up our observations on peer confrontation problems, we see that the dependency on the set $|\Omega|$ makes this kind of evaluation more problematic. Nonetheless, as AI research is becoming more socially oriented, with significantly more presence of multi-agent systems and game theory, an effort has to be done to make this kind of evaluation more systematic, instead of the plethora of arrangements that we see in sports, for instance. 

As a summary of this whole section about task-oriented evaluation, we have identified many issues in many evaluation settings in AI. Nonetheless, the three types of evaluation settings have their niches of application and task-specific evaluation is the right one in applications such as engineering, medicine, military devices, etc. In fact, the series of workshops on Performance Metrics for Intelligent Systems, held since 2000 at the National Institute of Standards \& Technology \cite{evans2001performance,meystel2003performance,madhavan2009performance,schlenoff2011performance} is a good example of the usefulness of this kind of evaluation. 
However, we hope that some ideas above can be used to make the evaluation more controlled, automated and robust.

\section{Towards ability-oriented evaluation}\label{sec:ability-oriented}

Many areas AI is successful nowadays took a long time to flourish in applications (e.g., driverless cars, machine translators, game bots, etc.). Most of them correspond to specific tasks and require task-oriented evaluation. Other tasks that are still not solved by AI technology are already evaluated in this way and will be successful one day. However, if instead of AI applications we think about AI systems, we see that there are some kinds of AI systems for which task-oriented evaluation is not appropriate. For instance, cognitive robots, artificial pets, assistants, avatars, smartbots, smart houses, etc., are not designed to cover one particular application but are expected to be customised by the user for a variety of tasks. In order to cover this wide range of (previously unseen) tasks, these systems must have some abilities such as reasoning skills, inductive learning abilities, verbal abilities, motion abilities, etc. Hence, this entails that apart from task-oriented evaluation methods we may also need ability-oriented evaluation techniques.

Things are more conspicuous when we look at the evaluation of the {\em progress} of AI as a discipline. If we look at AI with Minsky's 1968 definition seen in the introduction, i.e., by achievement of tasks that would require intelligence, AI has progressed very significantly. For instance, one way of evaluating AI progress is to look at a task and check in which category an AI system is placed: {\em optimal} if no other system can perform better, {\em strong super-human} if it performs better than all humans, {\em
super-human} if it performs better than most humans, {\em par-human} if it performs similarly to most humans, and {\em sub-human} if it performs worse than most humans \cite{rajani2011artificial}. Note that this approach does not imply that the task is necessarily evaluated with a human-discriminative approach. Having these categories in mind, we can see how AI has scaled up for many tasks, even before AI had a name. For instance, calculation became super-human in the nineteenth century, cryptography in the 1940s, simple games such as noughts and crosses became optimal in 1960s, more complex games (draughts, bridge) a couple of decades later, printed (non-distorted) character recognition in the 1970s, statistical inference in the 1990s, chess in the 1990s, speech recognition in the 2000s, and TV quizzes, driving a car, technical translation, Texas hold 'em poker in the 2010s. According to this evolution, the progress of AI has been impressive \cite{bostrom2014superintelligence}. The use of human intelligence as a baseline has been used in competitions (such as the humies awards\footnote{\url{www.human-competitive.org}.}) or to define ratios, where median human performance is set at a zero scale, such as the so-called Turing-ratio \cite{masum2002turing,masum2003turing}, with values greater than 0 for super-human performance and values lower than 0 for sub-human performance.

However, let us first realise that no system can do (or can learn to do) all of these things together. The big-switch approach may be useful for a few of them (e.g., a robot with an advanced computer vision system that detects whether it is facing a chess board or a bridge table and then switch to the appropriate program to play the game that it has just recognised). Second, if we look at AI with McCarthy's definition seen in the introduction, i.e., by making intelligent machines, things are less enthusiastic. Not only has the progress been more limited, but also there is a huge controversy for quantifying this progress (in fact, some argue that machines are more intelligent today than fifty years ago while others say that there has been no progress at all other than computational power). Hence, worse than having a poor progress or no progress at all,  we regard with contempt that we do not have effective evaluation mechanisms to evaluate this progress. It seems that none of the evaluation settings seen in the previous section is able to evaluate whether the AI systems of today are more intelligent than the AI systems of yore. Also, for developmental robotics and other areas of AI where systems are supposed to improve their performance with time, we want to know if a 6-month-old robot has progressed over its initial state, in the same way that we see how abilities increase and crystallise with humans, from toddlers to adults. Ability-oriented evaluation, and not task-oriented evaluation, seems to have a better chance of answering this question.

To make the point unequivocal, we could even go beyond McCarthy's definition of AI without the use of `intelligence' and define this view of AI as the {\em science and engineering of making machines do tasks they have never seen and have not been prepared for beforehand}. Clearly, this view puts more emphasis on learning, but it also makes it crystal clear that task-oriented evaluation, as have been performed for years, would not fit the above definition.

It would be unfair to forget to acknowledge that some attempts seen in the previous section have made an effort for a more general AI evaluation. The general game competition seen in the previous section is one example of how some things are changing in evaluation. Users and researchers are becoming tired of a big-switch approach. 
They yearn for and conceive systems that are able to cover more and more general task classes.  
Nonetheless, it is still a limited generalisation, which is too based on a very specific range of tasks. Many good players at the General Game Competition would be helpless at any game of the Arcade Learning Environments, and vice versa. Actually, only some reinforcement learning (and perhaps genetic programming) systems can at least participate in (adaptations to) of both games ---excelling in them would not be possible though without an important degree of specialisation.

In the rest of this section we will introduce what an ability is and how they can be evaluated in AI. The title of this section (starting with `Towards') suggests that what follows is more interdisciplinary and contains proposals that are not well consolidated yet, or that may even go in the wrong direction. Nonetheless, let us be more lenient and have in mind that ability-based evaluation is much more challenging than task-specific evaluation.

\subsection{\arxiv{What is an ability?}\journal{Cognitive abilities}}

We must first clarify that we are talking about {\em cognitive} abilities, in the same way that in the previous section we referred to {\em cognitive} tasks. Some AI applications require physical abilities, most especially in robotics, but AI deals with how the sensors and actuators are controlled, not about their strength, consumption, etc. After this clarification, we can define a cognitive ability as {\em a property of individuals that allows them to perform well in a range of information-processing tasks}. At first sight this definition may just look like a change of perspective (from problems to systems). However, what we see now is that the ability is required, and performance is worse without featuring the ability. In other words, the ability is necessary but it does not have to be sufficient (e.g., spatial abilities are necessary but not sufficient for driving a car). Also, the ability is assumed to be general, to cover a range of tasks. Actually, general intelligence would be one of these cognitive abilities, one that covers {\em all} cognitive tasks: ``general intelligence is a very broad trait that encompasses quickness and quality of response to all cognitive tasks'' \cite{strickler1973change}.

\begin{figure}[htdp]
	\centering
		\includegraphics[width=0.7\textwidth]{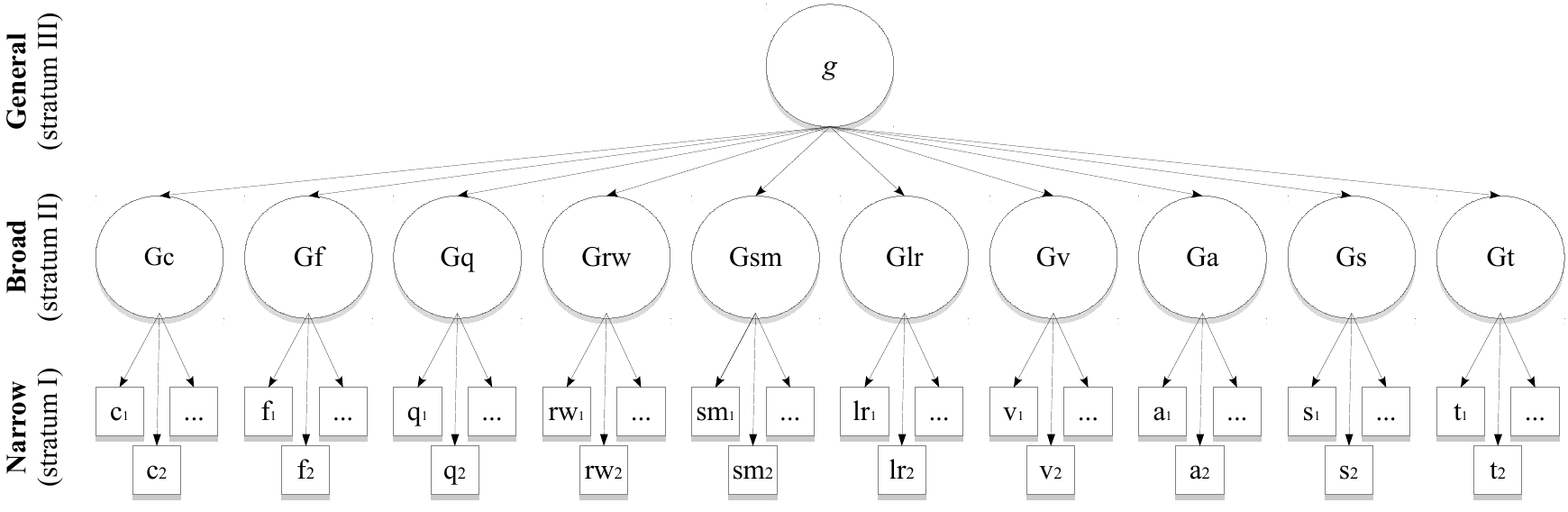}
	\caption{Cattell-Horn-Carroll's three stratum model.  
	The broad abilities are Crystallised Intelligence (Gc), Fluid Intelligence (Gf), Quantitative Reasoning (Gq), Reading and Writing Ability (Grw), Short-Term Memory (Gsm), Long-Term Storage and Retrieval (Glr), Visual Processing (Gv), Auditory Processing (Ga), Processing Speed (Gs) and 
Decision/Reaction Time/Speed (Gt).}
	\label{fig:factormodel}
\end{figure}

The major issue about abilities is that they are `properties', and as such they have to be conceptualised and identified. While tasks can be seen as measuring {\em instruments}, abilities are {\em constructs}. In psychology, 
many different cognitive abilities have been identified and have been arranged in different ways \cite{schaieprimary}. For instance, 
one well-known comprehensive theory of human cognitive abilities is 
the Cattell-Horn-Carroll theory \cite{keith2010cattell}. 
Figure \ref{fig:factormodel} shows a graphical representation of these abilities.  
The top level represents the {\em g} factor or general intelligence, the middle level identifies a set of broad abilities and the bottom level may include many narrow abilities. Again, this top level seems to saturate all tasks: ``g is common to all cognitive tasks including learning tasks''
\cite{alexander1997intelligence}.

  


Interestingly, this is not surprising from an AI standpoint. The broad abilities seem to correspond to subfields in AI. For instance, looking at any AI textbook (e.g., \cite{Russell-Norvig2009}), we can enumerate areas such as problem solving, use of knowledge, reasoning, learning, perception, natural language processing, etc., that would roughly correspond to some of the cognitive abilities in Figure \ref{fig:factormodel}.

Can we evaluate broad abilities as we did for specific tasks? Application-specific (task-oriented) approaches will not do. But is ability-oriented evaluation ready for this? The answer, as we will see below, is that this type of evaluation is still in a very incipient stage in AI. There are several reasons for this. First,  general (ability-oriented) evaluation is more challenging. Second, we no longer have a clear definition of the task(s). In fact, defining the ability depends on a conceptualisation, and from there we need to find a set of representative exercises that require the ability. And third, there have not been too many general AI systems to date, so task-oriented evaluation has seemed sufficient for the evaluation of AI systems so far. However, things are changing as new kinds of AI systems (e.g., developmental robotics) are becoming more general.

Before starting with some approaches in the direction of ability-oriented evaluation, it can be argued that some existing evaluation settings in AI are already ability-oriented. For example, even if the planning competition features a set of tasks, it goes around the {\em ability} of planning, which is more general than any particular task. However, the systems are not able to determine when planning is required for a range of problems. In other words, the ability is not a resource of the system, but the very goal of the system. In the end, it is the researchers who incorporate planning modules in several application-specific systems, and not the systems that independently enable their planning abilities to solve a new problem.

\subsection{The anthropocentric approach: psychometrics}\label{sec:psychometrics}

Psychometrics was developed by Galton, Binet, Spearman and many others at the end of the XIXth century and first half of the XXth century. An early concept that arose was the need of distinguishing tasks requiring very specific knowledge or skills from general abilities. For instance, an ``idiot savant'' could have a lot of knowledge or could have developed a sophisticated skill during the years for some specific domain, but could be obtuse for other problems. On the contrary a very able person with no previous knowledge could perform well in a range of tasks, provided they are culture-fair. This distinction took several decades to consolidate. In a way, this bears resemblance with the narrow vs. general dilemma in AI.

Psychometrics is concerned about measuring cognitive abilities, personality traits and other psychological properties \cite{Sternberg2000}. 
Factors differ from abilities, in principle, in that they are obtained through testing and further analysed through systematic approaches based on factor analysis. Some factors have been equated and named after existing abilities while others are `discovered' and receive new technical names. Several indices can be derived from a battery of tests by aggregating abilities and factors. One joint index that is usually determined from some of these tests is known as IQ (Intelligence Quotient). Although IQ was originally normalised by the subject's age (hence its name), its value for adults today is normalised relative to an adult population, assuming a normal distribution with mean $\mu$=100 and standard deviation $\sigma$=15. 
This corresponds to a more sophisticated (normalised) aggregation of results for several items, which again resembles our equation~\ref{eq:average}. 

IQ tests incorporate items of variable difficulty. Item difficulty is determined by the percentage of subjects that are able to solve the item, using functional models in Item Response Theory \cite{Lord80,embretson2000item}, as seen in the previous section. Note that this difficulty assessment is relative to the population and not derived from the nature of the item itself.

IQ tests are easy to administer, fast and accurate, and they are used by companies and governments, essential in education and pedagogy.
IQ tests are generally culture-fair through the use of abstract exercises (except for the verbal comprehension abilities).


As they work reasonably well for humans, their use for evaluating machines has been suggested many times, even since the early days of AI, with the goal of constructing ``a single program that would take a standard intelligence test'' \cite{newell1973you}. More recently, their use has been vindicated by 
Bringsjord and Schmimanski \cite{BringsjordSchimanski2003,bringsjord2011}, under the so-called `Psychometric AI' (PAI), as ``the {\em field} devoted to building information-processing entities capable of at least solid performance on all established, validated tests of intelligence and mental ability, a class of tests that includes not just the rather restrictive IQ tests, but also tests of artistic and literary creativity, mechanical ability, and so on''. 
It is important to clarify that PAI is a redefinition or new roadmap for AI ---not an evaluation methodology--- and does not further develop or adapt IQ tests for AI systems. In fact, PAI does not explicitly claim that IQ tests (or other psychometric tests) are the best way to evaluate AI systems, but it is said that an ``agent is intelligent if and only if it excels at all established, validated tests of intelligence'' (later broadened to any other psychometric test) \cite{BringsjordSchimanski2003,bringsjord2011}. The question of whether these tests are a necessary and sufficient condition for machines and the limitations of PAI as a guide for AI research have been recently discussed in \cite{besold2014}. 

Not surprisingly, this claim that IQ tests are the best way to evaluate AI system has recently come from human intelligence research. Detterman, editor of the Intelligence Journal, wrote an editorial \cite{Detterman2011} where he suggested that Watson (the then recent winner of the {\em Jeopardy!} TV quiz \cite{ferrucci2010building}) should be evaluated with IQ tests. The challenge is very explicit: ``I, the editorial board of {\em Intelligence}, and members of the International Society for Intelligence Research will develop a unique battery of intelligence tests that would be administered to that computer and would result in an actual IQ score'' \cite{Detterman2011}. Detterman established two levels for the challenge, a first level, where the type of IQ tests can be seen beforehand by the AI system programmer, and a second level, where the types of tests would have not seen beforehand. Only computers passing the second level ``could be said to be truly intelligent'' \cite{Detterman2011}. The need for two levels seems related to the big-switch approach and the problem overfitting issue, which we have already mentioned in previous sections for AI evaluation settings. It is apposite at this point to recall that academic and professional IQ tests and many other standardised psychological tests are never made public, because otherwise people could practise on them and game the evaluation. Note that the non-disclosure of the tests until evaluation time is something that we only find in very few evaluation settings in the previous section. 

Detterman was unaware that almost a decade before, in 2003, Sanghi and Dowe \cite{sanghidowe2003computer} implemented a small program (less than 1,000 lines of code) which could score relatively well on many IQ tests, as shown in Table~\ref{tab:sanghidowe}.
The program used a big-switch approach and was programmed to some specific kinds of IQ tests the authors had seen beforehand. The authors still made the point unequivocally: this program is not intelligent and can pass IQ tests.

\begin{table}
\begin{center}
\begin{tabular}{lcc}
\hline
Test                & IQ Score & Human Average \\ \hline
A.C.E. IQ Test	    & 108	   & 100     \\ 
Eysenck Test 1	    & 107.5	 & 90-110 \\
Eysenck Test 2	    & 107.5	 & 90-110 \\
Eysenck Test 3	    & 101	   & 90-110 \\
Eysenck Test 4	    & 103.25 & 90-110 \\
Eysenck Test 5	    & 107.5	 & 90-110 \\
Eysenck Test 6	    & 95	   & 90-110 \\
Eysenck Test 7	    & 112.5	 & 90-110 \\
Eysenck Test 8	    & 110	   & 90-110 \\
IQ Test Labs	      & 59	   & 80-120 \\
Testedich IQ Test   & 84	   & 100 \\
IQ Test from Norway & 60	   & 100 \\ \hline
Average	            & 96.27  & 92-108\\ \hline
\end{tabular}
\caption{Results by a rudimentary program for passing IQ tests (from \cite{sanghidowe2003computer}).}\label{tab:sanghidowe}
\end{center}
\end{table}

While it must be conceded that the results only reach the first level of Detterman's challenge ---so there is a test administration issue (i.e., an evaluation flaw)---, there are some weaknesses about human IQ tests that would also arise if a system passed the second level as well.
In particular, ``the editorial board of {\em Intelligence}, and members of the International Society for Intelligence Research'' could be tempted to devise or choose those IQ tests that are more `machine-unfriendly'. If AI systems eventually passed some of them, the battery could be refined again and again, in a similar way as how CAPTCHAs are updated when they become obsolete. In other words, this selection (or battery) of IQ tests would need to be changed and made more elaborate year after year as AI technology advances. Also, the limitations of this approach if AI systems ever become more intelligent than humans are notorious. 

The main problem about IQ tests is that they are anthropocentric, i.e., they have been devised for humans and take many things for granted. For instance, most assume that the subject can understand natural language to read the instructions of the exercise. 
On top of that, they are specialised to some human groups. For instance, tests are significantly different when evaluating small children, people with disabilities, etc.  Also, the relation between items and abilities have been studied during the past century exclusively using humans, so it is not clear that a set of items would measure the same ability for a human or for a machine. For instance, is it reasonable to expect that well-established tests of choice reaction time be correlated with intelligence in machines as they are correlated in humans \cite{deary2001reaction}? Or, what makes a set of psychometric tests different from a set of ``human intelligence tasks'' in Amazon Mechanical Turk \cite{buhrmester2011amazon}. For a more complete discussion about why IQ tests are not ready for AI evaluation, the reader is referred to a response \cite{IQnotformachines} to Detterman's editorial. 
%
%
Having said all this and despite the limitations of IQ tests for AI evaluation, their use is becoming more popular in the past decade (including robotics, \cite{schenck2013intelligence}) and systems whose results are like  those of Table \ref{tab:sanghidowe} are becoming common (for a survey, see \cite{computermodels-iq2014}, for an open library of IQ tests, see \url{pebl.sourceforge.net/battery.html}).

As just said, one of the problems of IQ tests is that they are specialised for humans. In fact, standardised adult IQ tests do not work with people with disabilities or children of different ages. In a similar way, we do not expect animals to behave well on a standard human IQ test, starting from the fact that they will not be able to read the text. This leads us to the consideration of how cognitive abilities are evaluated in animals. Comparative psychology and comparative cognition \cite{Shettleworth2010,shettleworth2013fundamentals} are the main disciplines that perform this evaluation. 
For a time, much research about cognitive abilities in animals was performed on apes. The term `chimpocentric' was introduced as a criticism about tests that had gone from being anthropocentric to being chimpocentric. Nonetheless, in the past decades, the perspective is much more general and any species may be a subject of study for comparative psychology: mammals (apes, cetaceans, dogs and mice), birds and some cephalopods. 
The evaluation focusses on ``basic processes'', such as perception, attention, memory, associative learning and the discrimination of concepts, and recently on more sophisticated instrumental or social abilities \cite{shettleworth2013fundamentals}.

One of the most distinctive features of animal evaluation is the use of rewards, as instructions cannot be used. This setting is very similar to the way reinforcement learning works. Animal evaluation has also brought attention to the relevance of the interface. Clearly, the same test may require very different interfaces for a dolphin and a bonobo. 

Human evaluation and animal evaluation have become more integrated in the past years, and testing procedures half way between psychometrics and comparative cognition are becoming more usual. For instance, several kinds of skills are evaluated in human children and apes in \cite{Herrmann-etal2007}. In recent years, many abilities that were considered exclusively human have been found to some extent in many animals.

Does the enlargement from humans to the whole animal kingdom suggest that these tests for animals can be used for machines? While the lower ranges of the studied abilities and the use of rewards can facilitate its application to AI systems significantly, we still have many issues about whether they can be applied to machines (at least directly). 
First, the selection of tasks and abilities is not systematic. Second, many of the tasks that are applied to animals would be too easy for machines (e.g., memory). 
And third, others would be too difficult (e.g., orientation, recognition and interaction in the real world). 
Nonetheless, there is an increasing need for the evaluation of animats \cite{williams2010information} and the evaluation procedures for animals are the first candidates to try.

\subsection{Evaluation using AIT}\label{sec:AIT}

A radically different approach to AI evaluation started in the late 1990s. If intelligence was viewed as a ``kind of information processing'' \cite{chandrasekaran1990kind} then it seemed reasonable to look at information theory for an ``essential nature or formal basis of intelligence and the proper theoretical framework for it'' \cite{chandrasekaran1990kind}. This was finally done with {\em algorithmic information theory} (AIT), and the related notions of Solomonoff universal probability \cite{solomonoff1964}, Kolmogorov complexity \cite{Li-Vitanyi08} and Wallace's Minimum Message Length (MML) \cite{Wallace-Boulton1968,Wallace-Dowe1999a}. 

There are several good properties about {\em algorithmic information theory} for evaluation. First, several definitions of information and complexity can be defined exclusively in computational terms, actually relative to a Universal Turing Machine (UTM), a fundamental and universal model of effective computation. For instance, the Kolmogorov complexity of an object (expressed as a binary string) relative to a UTM is defined as the shortest program (for that machine) that describes/outputs the object. Even if these definitions depend on the UTM that is used, the invariance theorem states that their values will only differ with respect to other UTM up to a constant that only depends on the two different UTMs (because one can emulate the other) \cite{Li-Vitanyi08}. The notion of algorithmic probability, introduced by Solomonoff, allows a universal distribution to be defined for each UTM, which is just the probability of objects as outputs of a UTM fed by a fair coin. While, in general, this means that compressible strings are more likely than incompressible ones, it can be shown that every computable probability distribution can be approximated by a universal distribution.  In a way, Solomonoff, the father of algorithmic probability \cite{solomonoff1964} gave a theoretical backing to Occam's razor.  There are reasons to think that many phenomena and, as a result, many of the problems that we face every day, follow a universal distribution. This is directly linked to equation~\ref{eq:average} again, and the discussion about the choice of the probability $p$. Also, we have the relevant fact, which is very significant for evaluation as well, that all universal distributions are immune to the no-free-lunch theorems, where system performance can differ very significantly for induction \cite{lattimore2013no,Hibbard2009}. And finally, Kolmogorov complexity and algorithmic probability are two sides of the same coin, which led to a formal connection 
of compression and inductive inference. It has been acknowledged that Solomonoff ``solved the problem of induction'' \cite{solomonoff1996does,David2013solomonoff}. 
	Of course, not everything in AIT is straightforward. For instance, some of these concepts lead to incomputable functions, although approximations exist, such as 
	Levin's $Kt$ \cite{Levin73}. 
	%

The application of AIT to (artificial) intelligence evaluation started with a variant of the Turing Test that featured compression problems \cite{Dowe-Hajek1997a,Dowe-Hajek1998} to make the test more sufficient. While one of the goals of this work was to criticise Searle's Chinese room \cite{Searle80} (an argument that has faded with time), this is one of the first intelligence test proposals using AIT. At roughly the same time, a formal definition of intelligence in the form of a so-called $C$-test was derived from AIT \cite{HernandezOrallo-MinayaCollado1998,HernandezOrallo2000a}. 
Figure \ref{fig:Ctest} shows examples of sequences that appear in this test. They clearly resemble some exercises found in IQ tests. The major differences are that (1) sequences are obtained by a generator (a UTM with some post-conditions about the generated sequence, ensuring the unquestionability of the series continuation and less dependency on the reference machine) and (2) the fact that each sequence is accompanied by a theoretical assessment of difficulty (a variant of Levin's $Kt$ complexity). Note the implications for evaluation of such a test, as exercises are derived from first principles (instead of being contrived by psychometricians) and the difficulty of these exercises is intrinsic, and not based on how difficult humans find them. Finally, these sequences were used to define a test by aggregating results 
in a way that highly resembles our recurrent equation~\ref{eq:average}, where $M$ is formally defined as including {\em all} possible sequences (following some conditions) and the probability is 
defined to cover a range 
of difficulties, leading to a difficulty-driven sampling as in Figure \ref{fig:sampling} (right).

\begin{figure}
\centering
{\sffamily\small

$$
\begin{array}{llll}
k=9  & : & $a, d, g, j, ...$ & $Answer: m$ \\
k=12 & : &  $a, a, z, c, y, e, x, ...$ & $Answer: g$ \\
k=14 & : & $c, a, b, d, b, c, c, e, c, d, ...$ & $Answer: d$  \\
\end{array}
$$
}
\vspace{-0.5cm}
\caption{Several series of different complexity 9, 12, and 14 used in the $C$-test \cite{HernandezOrallo2000a}.}
\label{fig:Ctest}
\end{figure}

Some preliminary experimental results showed that human performance correlated with the absolute difficulty ($k$) of each exercise and also with IQ test results for the same subjects. This encourages the use of this approach for IQ-test re-engineering. With the aim of a more complete test for machines, some extensions of the $C$-test were suggested, such as transforming it to work with interactive agents (``cognitive agents [...] with input/output devices for a complex environment'' \cite{HernandezOrallo-MinayaCollado1998} where ``rewards and penalties could be used instead'' \cite{HernandezOrallo00b}) or extending them for other cognitive abilities \cite{HernandezOrallo00d}. Despite its explanatory power about IQ tests, this line of research was sharply dashed in 2003 (at least as general intelligence tests for machines) by the evidence that very simple ---non-intelligent--- programs could pass IQ tests \cite{sanghidowe2003computer}, as we have discussed in section \ref{sec:psychometrics} (see Table \ref{tab:sanghidowe}). 

Nonetheless, the extension to interactive agents was performed anyway. 
Interestingly, when agents and environments are considered in terms of equation~\ref{eq:average}, we just find a performance aggregation over a set of environments, exactly as had been formulated several times in the past: 
``intelligence is the ability of a decision-making entity to achieve success in a variety of goals when faced with a range of environments'' 
\cite{fogel1991evolution}. Note that this roughly corresponds to the psychometric view of general intelligence as key to performance in a range (or all) cognitive tasks.  
%
A crucial aspect was then to define this {\em range} of environments, i.e., the choice of the distribution in equation~\ref{eq:average}.  
One option was to include {\em all} environments. In order to do this in a meaningful, elegant way (and get rid of any no-free lunch theorem), 
AIT and reinforcement learning were combined \cite{Legg-Hutter2007}. 
Equation~\ref{eq:average} was instantiated with all environments as tasks with a universal distribution for $p$, i.e., $p(\mu) = 2^{-K(\mu)}$, with $K(\mu)$ being the Kolmogorov complexity of each environment $\mu$. 
%
%
Another approach was to include all environments up to a given size or complexity, and a limit of steps \cite{dobrev2000,dobrev2005formal}. 

These proposals present several problems. First, some constructions are not computable, so approximations need to be used. 
Second, most environments are not really discriminative, and all agents will score the same, will just `die' or be stuck after a few steps (this issue is partially addressed, with the use of ergodic environments \cite{Legg-Hutter2007} or world(s) where agents cannot make fatal mistakes \cite{dobrev2000}). Third, overweighting very small environments (by the use of a universal distribution or a complexity limit) makes the definition very dependent on the reference machine chosen as environment generator.  Finally, time (or speed) is not considered for the environment or for the agent. For more details about these (and other) issues and some possible solutions, the reader is referred to \cite{Hibbard2009} and \cite[secs. 3.3 and 4]{HernandezOrallo-Dowe2010}. Taking into account these solutions, some actual tests have been developed \cite{CAEPIA2011, AGI2011Evaluating, AISB-AICAP2012a,leggveness2013approximation}. While the results may still be useful to rank some state-of-the-art machines, if they are not compared to humans (or animals), as we discuss in the following section, the validation (or more precisely the refutation) of these tests as true intelligence tests cannot be done.  

Summing up, the AIT approach is characterised by the definition of tests from formal information-based principles. This is in stark contrast to  other approaches where tasks are collected, refined by trial-and-error or invented in a more arbitrary way. Most of the approaches to AI evaluation using AIT seen above have aimed at defining and measuring {\em general} intelligence, which is placed at the very top of the hierarchy of abilities (and hence at the opposite extreme from a specialised task-oriented evaluation). However, many interesting things can happen if AIT is applied at other layers of the hierarchy, for general cognitive abilities other than intelligence, as suggested in \cite{HernandezOrallo00d} for the passive case and hinted in \cite[secs. 6.5 and 7.2]{HernandezOrallo-Dowe2010} for the dynamic cases, with the use of different kinds of videogames as environments (two of the most recently introduced benchmarks and competitions are in this direction \cite{bellemare13arcade,Schaul2014}). 
Finally, the information-theoretic approach is not isolated from some of the approaches seen so far in section \ref{sec:task-oriented}. Actually, some hybridisations and integrated approaches have been proposed \cite{AGI2011DarwinWallace,AGI2011Compression,AISB-AICAP2012b,manchester2012,AGI2012social} (apart from the compression-enriched Turing Tests (\cite{Dowe-Hajek1997a,Dowe-Hajek1998} already mentioned above).

\subsection{Universal psychometrics}

\arxiv{Figure~\ref{fig:fragmentation} shows the fragmentation of the approaches seen in previous sections. As we see, this} 
\journal{The previous sections show a fragmentation of techniques and problems. This} fragmentation is originated by the kind of measurement we are interested in (task-oriented or ability-oriented, collected or AIT-derived tests) but most especially by the kind of subject that is being measured. In \cite{HernandezOrallo-Dowe2010}, the notion of `universal test' is introduced, as a test that is applicable to ``any biological or artificial system that exists at this time or in the future'': human, non-human animal, enhanced human, machine, hybrid or collective. The stakes were set high, as the tests should work without knowledge about the subject, derive from computational principles, be unbiased (species, culture, language, ...), require no human intervention, be practical, produce a meaning score, and be anytime (the more time we have for the test the higher the reliability of the score). 
Note that in order to apply the same test to several subjects we are allowed to customise the interface, provided the features and difficulty of the items are remained unaltered. Also, we need to think about the speed of the subject, and adapt to it accordingly. Also, the capabilities of the subject can be quite varied, so the ranges of difficulty need to adapt to the agent. That suggests that universal tests must necessarily be adaptive.

\arxiv{
\begin{figure}[htdp]
	\centering
		\includegraphics[width=0.95\textwidth]{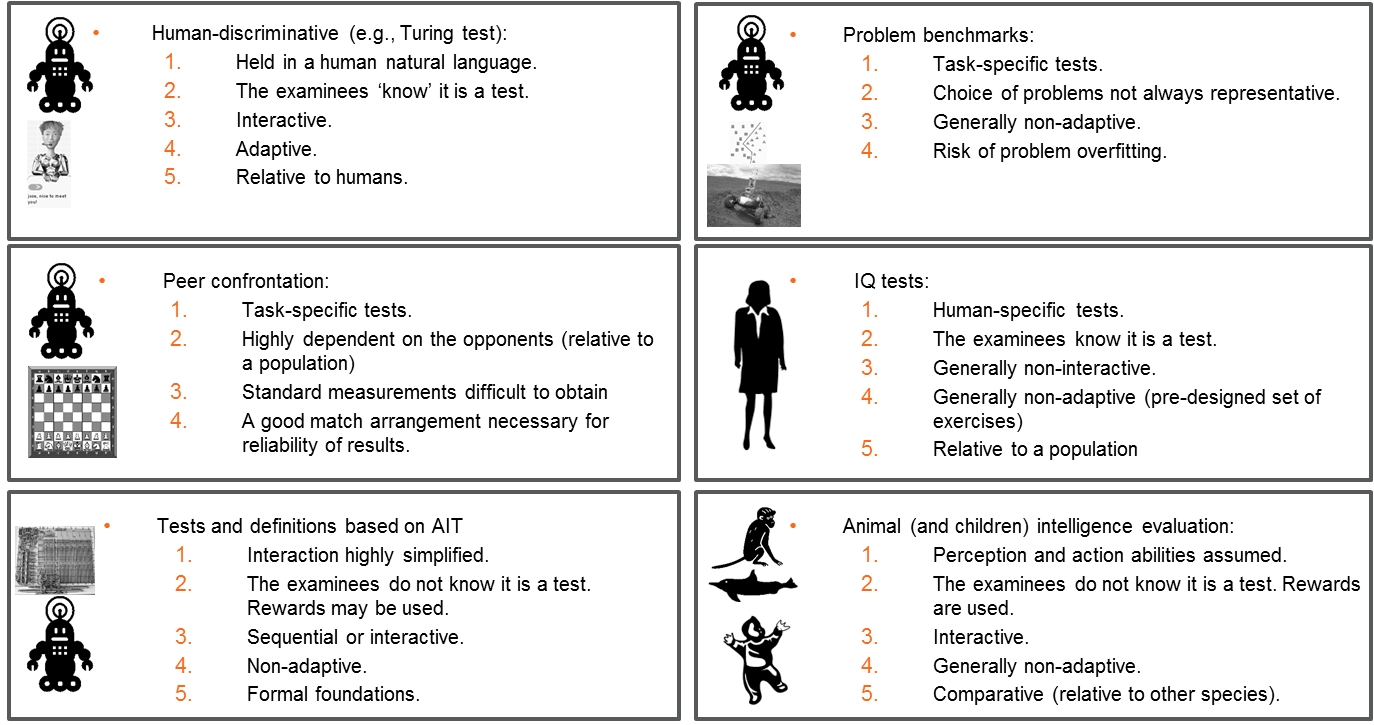}
	\caption{A schematic representation of the fragmentation of the different approaches for intellgience evaluation depending of the kind of intelligent systems. }
	\label{fig:fragmentation}
\end{figure}
}

A first framework for universal, anytime intelligence tests is introduced in \cite{HernandezOrallo-Dowe2010}, where a class of environment is carefully chosen to be discriminative. The test starts with very simple environments and adapts to the subject's performance and speed. In this regard, this resembles a difficulty-driven sampling as described in section \ref{sec:benchmarks}. The set of tasks (environments) were developed upon some of the ideas about using AIT for intelligence evaluation, as seen in section \ref{sec:AIT}. 
Some experiments were performed \cite{CAEPIA2011, AGI2011Evaluating, AISB-AICAP2012a}, using the environment class defined in \cite{HernandezOrallo10b}. Difficulty was estimated using a variant of Levin's $Kt$. As a way of checking whether the results were meaningful, the same test compared Q-learning \cite{watkins1992q} with humans. Two different interfaces were designed on purpose. 
The test gave consistent results for Q-learning and humans when considered separately, but were less reasonable when put together. 
The experimental settings featured many limitations (simplifications, non-adaptiveness, absence of noise, low-complexity patterns, no incrementality, no social behaviour, etc.) and, probably because of this, the results did not show the actual difference between Q-learning and humans. Despite the limited results, the experiment had quite a repercussion \cite{TheEconomist2011,NewScientist2011,TheConversation2011,yonck2012toward}.
Nonetheless, the tests were a first effort towards a universal test and highlighted some of the challenges.  

One concern about a generator of environments is the lack of richness of interaction and social behaviours that is expected. In other words, an environment that is randomly generated will have an extremely low probability of showing some social behaviour, which is a distinctive trait of human intelligence.  
This has suggested other ways of generating the environments and ways of incorporating other agents into them (e.g., the Darwin-Wallace distribution \cite{AGI2011DarwinWallace}), but  it is still an open research question how to adapt these ideas to the measurement of social intelligence and multi-agent systems \cite{AGI2011Compression,AGI2012social,orallo2014JAAMAS,insa2014}.

\begin{figure}
\vspace{-0.3cm}
\centering
\includegraphics[width=0.35\textwidth]{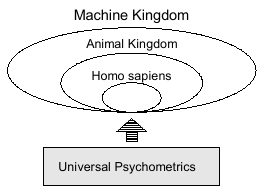}
\vspace{-0.2cm}
\caption{The realm of evaluable subjects for universal psychometrics.}
\label{fig:1kingdom}
\vspace{-0.3cm}
\end{figure}

\arxiv{The fragmentation of Figure~\ref{fig:fragmentation}}\journal{The fragmentation of approaches} and the need of solving many of the above issues has suggested the introduction of a new perspective, dubbed `universal psychometrics' \cite{upsychometrics2}. 
Universal psychometrics focusses on the measurement of cognitive abilities for the `machine kingdom', which comprises any (cognitive) system, individual or collective, either artificial, biological or hybrid. 
This comprehensive view is born with many hurdles ahead.  
Evaluation is always harder the less we know about the subject. The less we take for granted about the subjects the more difficult it is to construct a test for them.
For instance, human intelligence evaluation (psychometrics) works because it is highly specialised for humans. Similarly, animal testing works (relatively well) because tests are designed in a very specific way to each species. And some of the AI evaluation settings we have already seen work because they are specialised for some kind of AI systems that are designed for some specific applications. In the case of AI, who would try to tackle a more general problem (evaluating any system) instead of the actual problem (evaluating machines)?
The answer to this question is that the {\em actual} problem for AI is the {\em universal} problem. Notions such as `animat' \cite{williams2010information}, machine-enhanced humans \cite{cohentesting}, human-enhanced machines \cite{von2009human}, other kind of hybrids and, most especially, collectives \cite{quinn2011human} 
 of any of the former, suggest that the distinction between animals, humans and machines is not only inappropriate, but no longer useful to advance in the evaluation of cognitive abilities. The notion of `machine kingdom', as illustrated in Figure~\ref{fig:1kingdom}, is not very surprising to the current scientific paradigm but clarifies which class of subjects is most comprehensive.

Universal psychometrics attempts to integrate and standardise a series of concepts. A subject is seen as a physically computable (resource-bounded) interactive system. 
Cognitive tasks are seen as physically computable interactive systems with a score function. 
Interfaces are defined between subjects and tasks (observations-outputs, actions-inputs). 
Cognitive abilities are seen as properties over set of cognitive tasks (or task classes). As a result, the separation between task-specific and ability-specific becomes a progressive thing, depending on the generality of the class. 
Distributions are defined over task classes and results as aggregated performance on a task class (again, a generalised version of equation~\ref{eq:average}).
Difficulty functions are computationally defined from each task.
Overall, some of these elements found in psychometrics, comparative cognition and AI evaluation are overhauled here with the theory of computation and AIT. As a result, cognitive abilities are no longer {\em what the cognitive tests measure}, as in human psychometrics (adapting the (in)famous statement that intelligence is ``what intelligence test measures'' \cite{Boring1923}), but they are properties that emanate from (general) classes of tasks, perfectly defined in computational terms. As a consequence, the relation between abilities can be explored experimentally, but also theoretically, and measures are absolute and not relativised wrt. a population (except for social abilities). This could imply some revitalisation of the white-box approach, especially for those AI systems that can be formally described in a theoretical way (e.g., some results in \cite{hutter2007universal} and \cite{Hibbard2009} take a white-box evaluation approach).

This view of a cognitive ability is consistent with its association with a ``class of cognitive tasks'' \cite{carroll1993human} that must be `representative' for the ability. From the association between abilities and classes of tasks, ``we see that by merging two cognitive task classes we get a more general cognitive task class, and a more general ability. Typically, this is studied in a hierarchical way, starting with the so-called elementary cognitive tasks \cite[page 11]{carroll1993human} (closely related to the notion of primary mental abilities of \cite{thurstone1938primary})'' \cite{upsychometrics2}. This redraws our dilemma between task-oriented and ability-oriented into a gradual hierarchy from specific tasks to general abilities, with general intelligence at the very top (including {\em all} possible cognitive tasks, i.e., {\em all} interactive Turing machines with a score function). The questions about how to sample from a task class for an effective evaluation can be generalised from our discussion in section \ref{sec:task-oriented}.

This sets a dual view of cognitive tasks on one hand and cognitive systems on the other hand, where both spaces (the ability space and the machine kingdom) can be explored. Interestingly, both cognitive tasks and cognitive systems are defined as interactive systems, reflecting a duality world-agent. One singularity of cognitive systems (as well as their environments they are in) is that they can evolve with time, and their abilities can change. In other words, it seems that some abilities need to be constructed on top of other previously consolidated abilities, and this seems to be independent of the subject to some extent, in the same way that it seems difficult to be able to multiply without being able to add. A theoretical analysis of ability interdependency, how they can develop and the notion of potential intelligence, are still in a very incipient stage \cite{Hutter02,Hutter05,hernandez2013potential}.


There can be objections and disagreements about the way many of the above concepts should be understood and defined. 
There can also be objections about what a universal test should look like \cite{DoweHernandez-Orallo2013a_universal}. 
But a more integrated view of cognitive abilities for humans, animals, robots, agents, animats, hybrids, swarms, etc., is not only possible but useful. Bear in mind that universal psychometrics does not exclude the use of non-universal tests, as tests that are non-universal can be more efficient (tests can be universal or not, depending on the application), but aims at having a more integrated and well-founded view of how intelligent systems are evaluated in terms of cognitive abilities.

\section{Conclusions}\label{sec:conclusions}

\begin{figure}
\vspace{-0.3cm}
\centering
\includegraphics[width=0.4\textwidth]{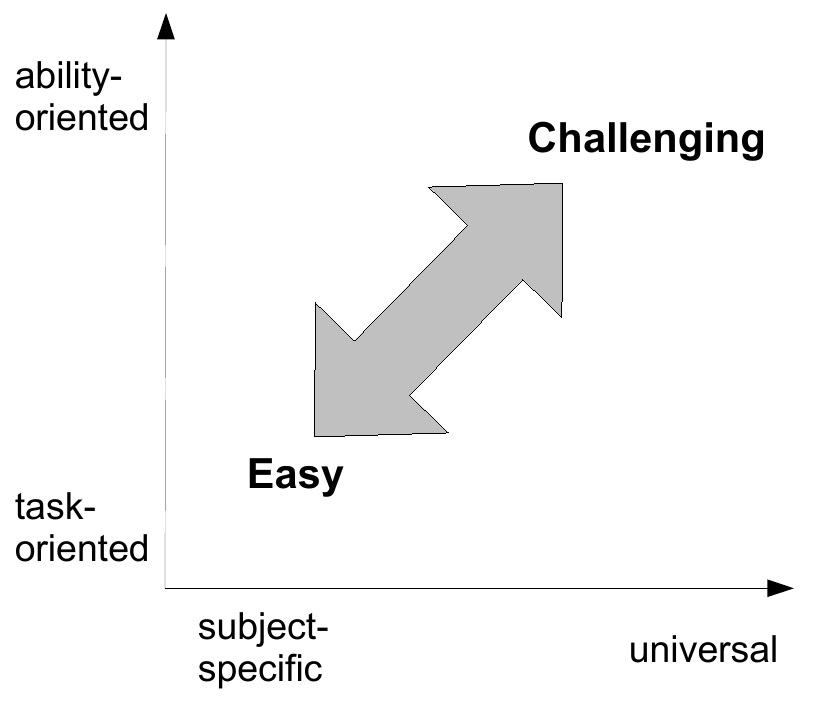}
\vspace{-0.2cm}
\caption{Tests become more or less challenging depending on the generality of the class of subjects under consideration (from subject-specific to universal) and the class of abilities (from task-oriented evaluation to ability-oriented evaluation).}
\label{fig:upsychometric-feasibility}
\vspace{-0.3cm}
\end{figure}

We started this paper looking at the way AI evaluation is commonly performed, through task-oriented evaluation, mostly with a black-box approach. We identified several problems and limitations, and we noticed that there is still a huge margin of improvement in the way AI systems are evaluated. The key issues are the set of tasks $M$ and their distribution $p$, as well as distinguishing the definition of the problem class (aggregation) from an effective sampling procedure (testing procedure). Then we switched to ability-oriented evaluation, a much more immature approach, but that may have a more relevant role in the future. The notion and evaluation of abilities is more elusive than the notion and evaluation of tasks. We have argued that this requires the integration of several perspectives that are currently scattered efforts in AI, psychometrics, AIT and comparative cognition. The different areas, philosophies, tools, foundations, terminologies and the different kinds of subjects to be evaluated can be unified with an integrated perspective known as universal psychometrics. Here, the exploration of the machine kingdom is dual to the exploration of the set of possible cognitive abilities/tasks. In both spaces we aim at becoming more general, which is where evaluation is more challenging (see Figure~\ref{fig:upsychometric-feasibility}). This resembles the duality in the theory of computation (e.g., problem classes and automata classes). The more formal approach advocated by universal psychometrics can make the white-box evaluation approach recover some relevance in AI.

From the problems and limitations found in AI evaluation and the tools and ideas that have appeared along the paper, we now enumerate a number of generic guidelines. These can be considered when an AI evaluation setting is under consideration.

\begin{itemize}
\item The definition of $\Omega$, the set of possible systems that can be evaluated (or that can be opponents in peer confrontation evaluation), must be clarified from the beginning. Any information about their proficiency and expected characteristics may be very useful. If humans are considered, the way in which they are admitted and how they are instructed must be defined. The more general $\Omega$ is the less we can assume about the evaluation process. If $\Omega$ is heterogeneous (e.g., a universal test), different interfaces must be considered.
\item The definition of $M$, the set of possible tasks, and its associated distribution $p$ configure what we are measuring. This can be built from a set of problems or using a generator. This pair $\left\langle M, p \right\rangle$  has to be representative of a task (in task-oriented evaluation) or an ability (in ability-oriented evaluation). If it is a peer confrontation evaluation, $M$ will be enlarged with as many combinations between game (environment) and agents in $\Omega$ are possible. The distribution $p$ will be updated accordingly. 
\item The definition of $R$ and its aggregation $\Phi$ must ensure that the values $R(\mu)$ for all $\mu \in M$ are going to be commensurate and that the aggregation is bounded. 
An analysis about expected measurement error is useful at this point. The robustness of  $R$ depending on the length or time left for each episode will indicate whether repetitions are needed to reduce the measurement error given by $R(\mu)$.
\item As much as possible, the similarity between tasks or a set of features describing them should be identified. An intrinsic difficulty function (even if approximate) is always very useful. Showing the distribution of difficulty for $M$ can be highly informative. If difficulty is available, item response curves could be prepared. 
\item The sampling method must be as much efficient as possible, by using, e.g., a clustering sampling or a range of difficulties if we have a non-adaptive 
evaluation. For the peer-confrontation evaluation, the arrangement of matches can be designed beforehand if the evaluation is not adaptive. Similarly, the procedure for an adaptive evaluation must also be carefully designed to ensure measurement robustness. Simulations can be useful to estimate this.
\item Information about how the evaluation is performed (including $R$, $\Phi$ and some illustrative problems) can be disclosed to the systems that are being evaluated (or to their designers). However, $\Omega$, $M$ and $p$ should not be disclosed. If possible, the problems should not be disclosed after the evaluation either, as keeping them secret makes it possible to compare with the same problems for different subjects or at different times (e.g., we can evaluate progress of a system or a discipline during a period).
\item After the evaluation, results must be analysed beyond the mere calculation of the aggregated results. Item response functions and agent response functions \cite{upsychometrics2} can be constructed empirically from the results and compared with the theoretical functions or any other information about $\Omega$ and $M$. Discrepancies or anomalies may suggest that the evaluation setting has to be revised. Results of the evaluation must become public at the highest possible detail, so they can be analysed and compared by other researchers and participants (following, e.g., the notion of `experiment database' \cite{vanschoren2012experiment}, such as in the machine learning community\footnote{\url{http://openml.org/}.}). 
\end{itemize}

\noindent It is of course an open question to what extent the above recommendations will be followed on a regular basis for AI evaluation. 
It can be argued whether AI evaluation has been a priority for AI in the past, but it seems that it has not been recognised as an imperative problem or a mainstream area of research. If this is the case, this paper can help change this perspective. Anyhow, the question of AI evaluation remains and there is space for significant improvement, even for the most specific sets $\Omega$ and $M$ (bottom-left part of Figure~\ref{fig:upsychometric-feasibility}). At the other end, measuring intelligence and doing it universally is a key ingredient for understanding what intelligence is (and, of course, to devise intelligent artefacts). Many interesting questions and applications lay in the middle of Figure~\ref{fig:upsychometric-feasibility}, as AI evaluation is no longer limited to task-specific evaluation of AI systems or to evaluating progress in AI. 
Instead, AI is becoming able to evaluate systems that learn to solve instead of systems that are programmed to solve. 

In any case, and with any of the approaches seen so far, a more scientific theory of AI evaluation is being required for many applications (CAPTCHAs, social networks, agent certification, etc.) and it will be more and more common in a future with a plethora of bots, robots, artificial agents, avatars, control systems, `animats', hybrids, collectives, etc. It is also crucial for the technological singularity once (and if) achieved \cite{eden2013singularity}, especially because some of the prophecies and forecasts disregard that the first thing to consider about the singularity is to have metrics to detect whether and where AI progresses towards it.

Summing up, AI requires an accurate, effective, non-anthropocentric, meaningful and computational way of evaluating its progress, by evaluating its artefacts. This paper can serve as a comprehensive source of the state of the art of the AI evaluation, its challenges and the avenues for future work.

\arxiv{
\section*{Acknowledgements}
%
{\small 
This work was supported by the EU (FEDER) and the Spanish MINECO projects CONSOLIDER-INGENIO CSD2007-00022,  TIN 2010-21062-C02-02, and TIN 2013-45732-C4-1-P, by Generalitat Valenciana projects Prometeo/2008/051 and PROMETEO/2011/052, and the REFRAME project granted by the European Coordinated Research on Long-term Challenges in Information and Communication Sciences \& Technologies ERA-Net (CHIST-ERA), and funded by 
Ministerio de Econom\'{\i}a y Competitividad, with code PCIN-2013-037.
%
I thank the organisers of the Summer School of the Spanish Association for Artificial Intelligence, in A Coru\~na, Spain, held in September 2014, for giving me the opportunity to give a lecture on `AI Evaluation'. This paper evolved in parallel with that lecture. The coverage of the BotPrize competition was discussed with Manuel Gonz\'alez-Bedia. Figure~\ref{fig:factormodel} is courtesy of Fernando Mart\'{\i}nez-Plumed.
}
}

{\small
\bibliography{biblio}
}

\end{document}